\documentclass{article}

\PassOptionsToPackage{numbers}{natbib}
\usepackage{amssymb}
\usepackage{amsmath,amsfonts}
\usepackage{amsopn}
\usepackage{bm} 
\usepackage{multirow}
\usepackage{algorithm}
\usepackage{algorithmic}
\usepackage{textcomp}
\usepackage{csquotes}
\usepackage{graphicx}
\usepackage{booktabs}
\usepackage{hyperref}
\usepackage{amsmath}
\usepackage{amssymb}
\usepackage{mathtools}

\usepackage{amsthm}
\usepackage[capitalize,noabbrev]{cleveref}
\usepackage[title]{appendix}
\usepackage{subcaption}
\usepackage{wrapfig}
\usepackage[utf8]{inputenc} 
\usepackage[T1]{fontenc}    
\usepackage{url}            
\usepackage{xspace}
\usepackage[export]{adjustbox}
\usepackage{amsfonts}       
\usepackage{nicefrac}       
\usepackage{microtype}      
\usepackage{wasysym}
\usepackage{enumitem}
\usepackage[american]{babel}
\usepackage{multirow}
\usepackage{adjustbox}
\usepackage[utf8]{inputenc} 
\usepackage[T1]{fontenc}    
\usepackage{hyperref}       
\usepackage{url}            
\usepackage{booktabs}       
\usepackage{amsfonts}       
\usepackage{nicefrac}       
\usepackage{microtype}      
\usepackage{xcolor}         
\usepackage{pifont}
\usepackage{enumitem}
\usepackage{listings}

\newcommand{\vct}[1]{\boldsymbol{#1}} 

\usepackage[preprint]{Neurips2025/neurips_2025}

\usepackage[utf8]{inputenc} 
\usepackage[T1]{fontenc}    
\usepackage{hyperref}       
\usepackage{url}            
\usepackage{booktabs}       
\usepackage{amsfonts}       
\usepackage{nicefrac}       
\usepackage{microtype}      
\usepackage{xcolor}         

\newcommand{\ProbOpr}[1]{\mathbb{#1}}

\newcommand{\expect}[2]{%
\ifthenelse{\equal{#2}{}}{\ProbOpr{E}_{#1}}
{\ifthenelse{\equal{#1}{}}{\ProbOpr{E}\left[#2\right]}{\ProbOpr{E}_{#1}\left[#2\right]}}} 

\newcommand{\x}{{\vct{x}}}

\newcommand{\sD}{\mathcal{D}}
\newcommand{\sM}{\mathcal{M}}

\newcommand{\sV}{\mathcal{V}}
\newcommand{\eat}[1]{}

\newcommand{\bbR}{\mathbb{R}}

\definecolor{darkgreen}{rgb}{0.0, 0.4, 0.26}  
\definecolor{darkred}{rgb}{0.7, 0.15, 0.15}   
\newcommand{\name}{CoT$^2$\xspace}

\title{Make Still Further Progress: \\ Chain of Thoughts for Tabular Data Leaderboard}

\author{Si-Yang Liu\textsuperscript{1,2}\,\textsuperscript{*} ~~~
  Qile Zhou\textsuperscript{1,2}~\footnotemark[1] ~~~
  Han-Jia Ye\textsuperscript{1,2} \\
  \\
  $^1$ School of Artificial Intelligence, Nanjing University, China \\
  $^2$ National Key Laboratory for Novel Software Technology, Nanjing University \\
}

\begin{document}

\maketitle
\let\thefootnote\relax\footnotetext{* Equal contribution.}

\begin{abstract}
Tabular data, a fundamental data format in machine learning, is predominantly utilized in competitions and real-world applications. The performance of tabular models---such as gradient boosted decision trees and neural networks---can vary significantly across datasets due to differences in feature distributions and task characteristics. Achieving top performance on each dataset often requires specialized expert knowledge.
To address this variability, practitioners often aggregate the predictions of multiple models. However, conventional aggregation strategies typically rely on static combination rules and lack instance-level adaptability.
In this work, we propose an in-context ensemble framework for tabular prediction that leverages large language models (LLMs) to perform dynamic, instance-specific integration of external model predictions.  
Without access to raw tabular features or semantic information, our method constructs a context around each test instance using its nearest neighbors and the predictions from a pool of external models. 
Within this enriched context, we introduce Chain of Tabular Thoughts (CoT$^2$), a prompting strategy that guides LLMs through multi-step, interpretable reasoning, making still further progress toward expert-level decision-making.
Experimental results show that our method outperforms well-tuned baselines and standard ensemble techniques across a wide range of tabular datasets.\looseness=-1
\end{abstract}

\section{Introduction}
\label{sec:intro}
Tabular data holds a pivotal position in the field of machine learning, primarily because of its organized and accessible format. In many competitions and real-world applications, ranging from financial forecasting~\cite{addo2018credit} to healthcare diagnostics~\cite{HassanAHK20}, tabular data serves as the primary data format. 
Recently, Gradient Boosted Decision Trees (GBDTs)~\cite{chen2016xgboost,ke2017lightgbm, Prokhorenkova2018Catboost} and Neural Networks (NN)~\cite{GorishniyRKB21Revisiting,ye2023ptarl,borisov2022deep} are two of the most commonly explored methods for tabular data learning~\cite{jiang2025representationlearningtabulardata}. However, although GBDTs often outperform NNs across many datasets~\cite{Grinsztajn2022Why}, the diverse nature of tabular data tasks implies that either method could be the most or least effective choice for a specific dataset~\cite{McElfreshKVCRGW23when,Ye2024Closer}. 
In practice, achieving high accuracy often requires expert-level tuning and the integration of multiple models. For instance, top solutions in machine learning competitions frequently adopt ensemble strategies designed by experienced practitioners.

Large Language Models (LLMs)\cite{achiam2023gpt,brown2020language} have achieved remarkable success across a range of domains, including question answering\cite{yu2023natural}, code generation~\cite{OpenCodeInterpreter}, and scientific reasoning~\cite{OminiScience}. However, LLMs' application to tabular data prediction remains limited. This is primarily because tabular tasks often involve reasoning over numerical values~\cite{DBLP:conf/nips/ManikandanJK23}—such as predicting house prices based on features like square footage and number of bedrooms—where LLMs typically underperform.
Current research on applying LLMs to tabular data remains limited and is mostly constrained to datasets with comprehensive textual descriptions. Existing studies can be broadly categorized into two main approaches:
One line of work directly converts tabular instances into text prompts using feature descriptions, allowing the LLM to act as a predictor~\cite{dinh2022lift}. This approach is heavily dependent on the availability and quality of textual descriptions and often struggles with numerical precision.
The other line of research uses LLMs to support traditional tabular pipelines by automating steps like data cleaning~\cite{LLMCleaning}, feature engineering~\cite{hollmann2024large}, and hyperparameter tuning~\cite{MLCopilot}. 
However, the effectiveness of these methods remains fundamentally constrained by the richness and accessibility of semantic information. In many practical scenarios—especially those involving sensitive data or proprietary systems—such semantic information, including feature names or task-level descriptions, may be unavailable or inaccessible. This limitation hinders the deployment of LLM-based methods that rely on explicit textual representations. Motivated by this, 
we pose the following question: 

\begin{displayquote}
\textit{
\noindent When there are no textual descriptions, can we transform the LLM into a competition expert, leveraging its robust reasoning abilities to make predictions with minimal computational cost?
}
\end{displayquote}

\begin{wrapfigure}{r}{0.5\textwidth} 
\includegraphics[width=\linewidth]{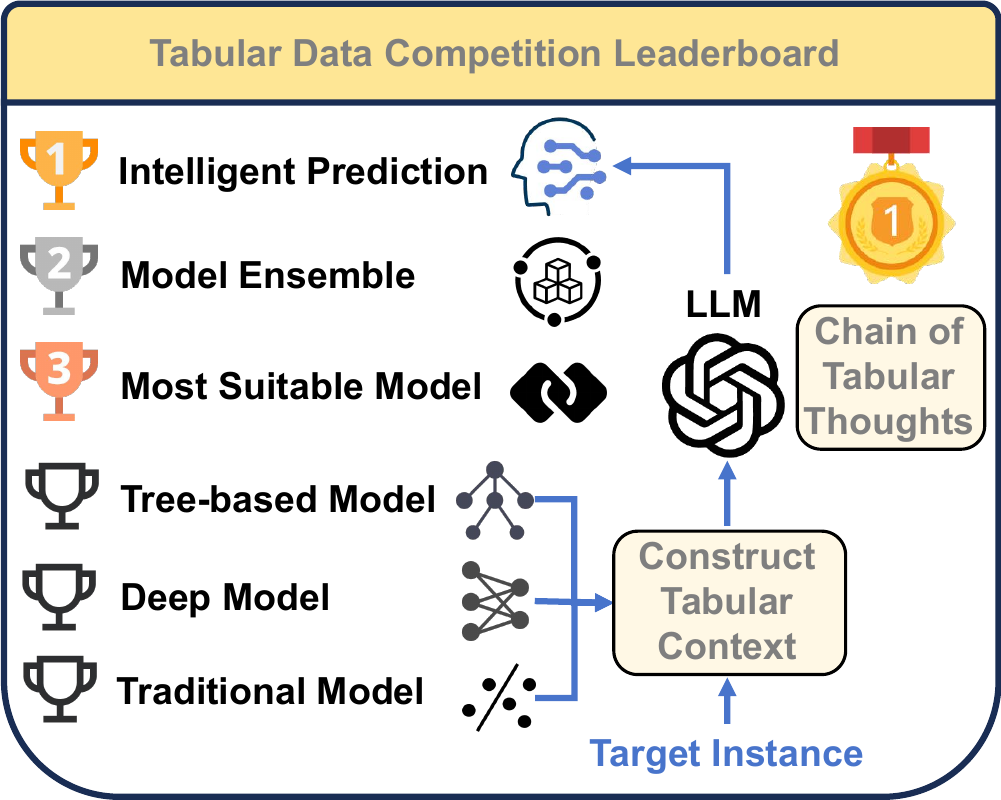}
\centering
  \caption{\name utilizes the expert knowledge of LLMs to create an intelligent ensemble of tabular models, making still further progress.}
\label{fig:leaderboard}
\end{wrapfigure}

To address the question above, our central idea is to empower LLMs to act like human experts in machine learning competitions: rather than directly accessing raw features, the LLM integrates multiple model predictions at the instance level, forming a dynamic ensemble guided by contextual knowledge. This approach leverages the LLMs’ general reasoning capabilities to selectively synthesize and reconcile external predictions, much like how practitioners deliberate over conflicting model outputs when making final decisions.
To realize this goal, we identify three key challenges:

First, \textit{where does the knowledge come from}? In many real-world scenarios, the semantic richness of tabular data is often limited. For instance, features may consist solely of numerical outputs from multiple sensors, or data privacy concerns may restrict access to descriptive information. Thus, a critical objective is to construct a context that conveys essential predictive signals while avoiding reliance on raw features or textual descriptions. Unlike domains where data is inherently sequential and context-rich, tabular data lacks natural contextual structure. To overcome this, we construct a tabular context for each target instance. We first identify local neighbors of the target and gather predictions from multiple external models within this neighborhood, combined with other non-semantic dataset information. This synthesized context is then used as a prompt to the LLM, which generates a final prediction. However, our empirical analysis shows that LLMs do not inherently interpret such context effectively without further guidance.

Second, \textit{how do we guide the LLM to think}?
Simply exposing the LLM to tabular contexts is not enough—we need to guide it to reason like an expert. Inspired by the Chain of Thought (CoT)~\cite{wei2022chain}, we introduce a structured reasoning process tailored for tabular data: the Chain of Tabular Thoughts (\name).~\name decomposes the prediction process into multiple analytical steps, such as identifying outliers and selecting appropriate models, leveraging the interactions among neighboring instances and their associated predictions from external models. By guiding the LLM through this step-by-step reasoning, we enable it to detect anomalies and select the best models for the local neighborhood.~\name compensates for the LLMs’ limited sensitivity to raw numerical values, helping it make effective and interpretable predictions like a machine learning competition expert~\autoref{fig:leaderboard}.

After equipping LLMs with carefully constructed tabular contexts and chains of thought, the third challenge is \textit{minimizing inference cost}. Since LLM-based methods require running inference for each target instance~\cite{dinh2022lift, hegselmann2023tabllm, tabula8b}, reducing the number of instances that invoke LLMs is crucial for practical deployment. In real-world settings, many instances are relatively simple---multiple external models already yield consistent and accurate predictions for them. We identify such cases by measuring the agreement among external model outputs and bypass LLM processing when sufficient consensus is observed. As a result, LLMs are reserved for more challenging instances, where model disagreement indicates greater uncertainty. This selective strategy significantly reduces computational overhead while preserving the benefits of LLM reasoning for complex cases.

We validate the effectiveness of our method on the TinyBench2 benchmark~\cite{Ye2024Closer}, which surpasses ensemble methods and well-tuned baselines, making further progress on the leaderboard. 
In summary, our main contributions are as follows:
\begin{itemize}[noitemsep,topsep=0pt,leftmargin=*]
\item We propose a novel tabular context construction method that removes the reliance of LLMs on textual datasets or feature descriptions, thereby significantly enhancing the applicability and privacy-preserving potential of LLMs in tabular domains.
\item We present the \textit{Chain of Tabular Thoughts }(\name) approach, which enables step-by-step reasoning and decision-making, effectively unlocking the numerical and logical reasoning capabilities of LLMs on tabular data.
\item We are the first to explore the role of LLMs in model ensembling for tabular prediction, addressing a previously overlooked yet crucial component in the modeling pipeline, and extending the use of LLMs beyond existing applications such as feature engineering or data cleaning.
\end{itemize}

\section{Related Work}

\subsection{Tabular Data Learning}
 Tabular data is among the most prevalent data formats in real-world machine learning applications. Gradient-boosted decision trees (GBDTs)~\cite{chen2016xgboost,Prokhorenkova2018Catboost,ke2017lightgbm} remain a dominant and highly competitive approach for tabular prediction tasks due to their efficiency and strong empirical performance. As ensemble-based models, GBDTs iteratively construct decision trees to minimize residual loss, making them well-suited for capturing heterogeneous patterns common in tabular datasets~\cite{Rubachev2024TabRed,cai2025understanding}.
Meanwhile, the rapid development of deep learning has led to a surge of interest in adapting neural architectures for tabular data~\cite{borisov2022deep}. These efforts include MLP-based variants~\cite{KlambauerUMH17SNN,GorishniyRKB21Revisiting,David2024RealMLP}, architectures tailored for tabular structures~\cite{WangFFW17DCN,Chen2023TabCaps}, attention-based models~\cite{Huang2020TabTransformer,Chen2023Excel,Zhou2023TabToken,JiangYW00Z24Tabular}, regularization-enhanced frameworks~\cite{ye2023ptarl,Wu2024SwitchTab}, and tree-inspired~\cite{ArikP21TabNet,Badirli2020GrowNet,PopovMB20Neural,zhou2019deep} or context-aware methods~\cite{gorishniy2023tabr,Ye2024ModernNCA}. Despite these innovations, recent large-scale benchmarks~\cite{Grinsztajn2022Why,Ye2024Closer,McElfreshKVCRGW23when} consistently show that GBDTs still outperform deep models in most tabular tasks. While several deep learning methods have attempted to mimic ensembling effects~\cite{PopovMB20Neural,Badirli2020GrowNet,DBLP:conf/icml/Chen2023Trompt}, few have succeeded in consistently closing the gap. Recent advances such as TabM~\cite{Yury2024TabM} and {\sc Beta}~\cite{Liu2025Beta}, which integrates BatchEnsemble~\cite{WenTB20BatchEnsemble} into tabular networks, show that efficient and scalable ensembling in deep tabular models remains an active and promising direction.

\subsection{\bf Language Models for Tabular Prediction}
\label{sec: LLM-method}
Although Pre-trained Language Models have achieved success in various fields on unseen tasks, their application to tabular data is often limited due to the prevalence of numerical values and the scarcity of textual descriptions. Additionally, concerns over data privacy and security can further restrict the availability of semantic information. As a result, the use of language models in tabular datasets is typically confined to scenarios where textual data is sufficient. 
TransTab~\cite{Wang2022TransTab} trains a tokenizer based on the words present in the tabular data to aid in prediction, rather than using a language model directly. TP-BERTa \cite{yan2024making} does not choose large language models. It fine-tunes relatively smaller pre-trained language models such as RoBERTa~\cite{liu2019roberta} for tabular data prediction. Some other methods start by serializing data through feature names into text, combining this with task descriptions to enable direct predictions by LLMs~\cite{dinh2022lift,hegselmann2023tabllm,tabula8b}. 
Among them, LIFT~\cite{dinh2022lift} requires fine-tuning on the whole training set, while TabuLa-8B~\cite{tabula8b} and TabLLM~\cite{hegselmann2023tabllm} focuses on data scarce scenarios.

\subsection{Retrieval-augmented Generation}
Retrieval-Augmented Generation (RAG) was originally developed in the language modeling domain to address the limitations of LLMs on knowledge-intensive tasks~\cite{lewis2020RAG,gao2023SurveyRAG}, enabling models to incorporate external knowledge bases for more accurate and informed responses. However, the use of RAG in tabular data learning remains relatively limited. A notable exception is TabR~\cite{gorishniy2023tabr}, which retrieves nearest neighbors to enhance neural tabular model representations.  
Recent studies such as LocalPFN~\cite{Thomas2024LocalPFN} and TabDPT~\cite{MaTabDPT} further demonstrate that leveraging local neighbors to construct context significantly enhances the performance of tabular foundation models (\textit{e.g.} TabPFN~\cite{hollmann2025TabPFNv2}, TabICL~\cite{TabICL_foundation_model}, and TabPTM~\cite{Ye2023TabPTM}). These approaches suggest that incorporating instance-specific, retrieval-based context not only improves generalization but also facilitates more efficient adaptation to downstream tasks~\cite{SF_PFN,koshil2024towards,ye2025closer}. 
This retrieval-based paradigm has also been extended to enhance tabular prediction with LLMs. \cite{TabICL_LLM} applies the RAG mechanism to enable large language models to effectively process large-scale tabular datasets, constructing informative contexts through instance-level neighbor retrieval.
In our approach, we use the labels of retrieved neighbors and the prediction outputs of external models as key components of the context for~\name's reasoning. However, instead of relying on the LLM to directly perform classification or regression, we position it as an intelligent ensembling agent. This design allows the LLM to make informed decisions by reasoning over the structured outputs, without accessing any raw tabular features or semantic information. As a result, our method offers strong privacy protection while retaining the benefits of instance-aware, context-driven prediction.

\subsection{LLMs for Enhancing Machine Learning Pipelines.}
Despite the success of machine learning (ML) in real-world tasks, building effective ML pipelines remains challenging due to the many design choices involved. AutoML~\cite{hutter2019automated} aims to automate this process through methods such as neural architecture search~\cite{pham2018efficient} and Bayesian optimization~\cite{frazier2018tutorial}. While effective, most AutoML techniques are time-consuming, lack transferability across tasks, and often behave as black boxes with limited interpretability~\cite{MLCopilot}.
To overcome these challenges, recent efforts have explored using Large Language Models (LLMs) to enhance ML workflows. LLM-based agents can assist with various stages of the pipeline—including task understanding~\cite{HardML,chan2025mlebench,infiAgent}, data cleaning~\cite{LLMCleaning,bodensohn2025unveilingchallengesllmsenterprise}, feature engineering~\cite{hollmann2024large,OCTree,LLMSimpleFeature,han2024large,DBLP:conf/bigdataconf/ZhangL24}, and model building and tuning~\cite{AutoKaggle,MLCopilot,AutoMLGPT,MLAgentBench,AIDE}—but most of these methods depend heavily on semantic information such as column descriptions or dataset metadata. Notably, no prior work has explored using LLMs as intelligent ensemble experts for tabular prediction tasks.
Our approach addresses this gap by treating the LLM not as a direct predictor, but as an instance-aware decision-maker that integrates outputs from multiple external models and nearest-neighbor labels. This enables accurate, interpretable predictions without accessing raw features or semantic cues, thus preserving privacy while enhancing performance.

Our method targets a fundamentally different setting from prior LLM-based approaches for tabular data. Existing methods largely fall into two categories: (1) approaches that convert each instance into a textual prompt using feature names or dataset descriptions, allowing the LLM to act as a predictor~\cite{dinh2022lift,hegselmann2023tabllm,tabula8b}; and (2) LLM-assisted tools that help automate parts of the ML pipeline—such as data cleaning, feature engineering, or hyperparameter tuning—which also rely heavily on task instructions or column-level semantics~\cite{LLMCleaning,hollmann2024large,MLCopilot}. In contrast, our method assumes no access to raw features or semantic descriptions. Instead, we position the LLM as an instance-wise ensemble expert that reasons over structured outputs (e.g., model predictions and neighbor labels), enabling accurate and interpretable predictions even in privacy-sensitive or low-semantic settings. Owing to this distinct problem formulation, these existing approaches fall outside the scope of our empirical comparisons.
\section{Methods}
\begin{figure}[t]
  \centering
  \includegraphics[width=\linewidth]{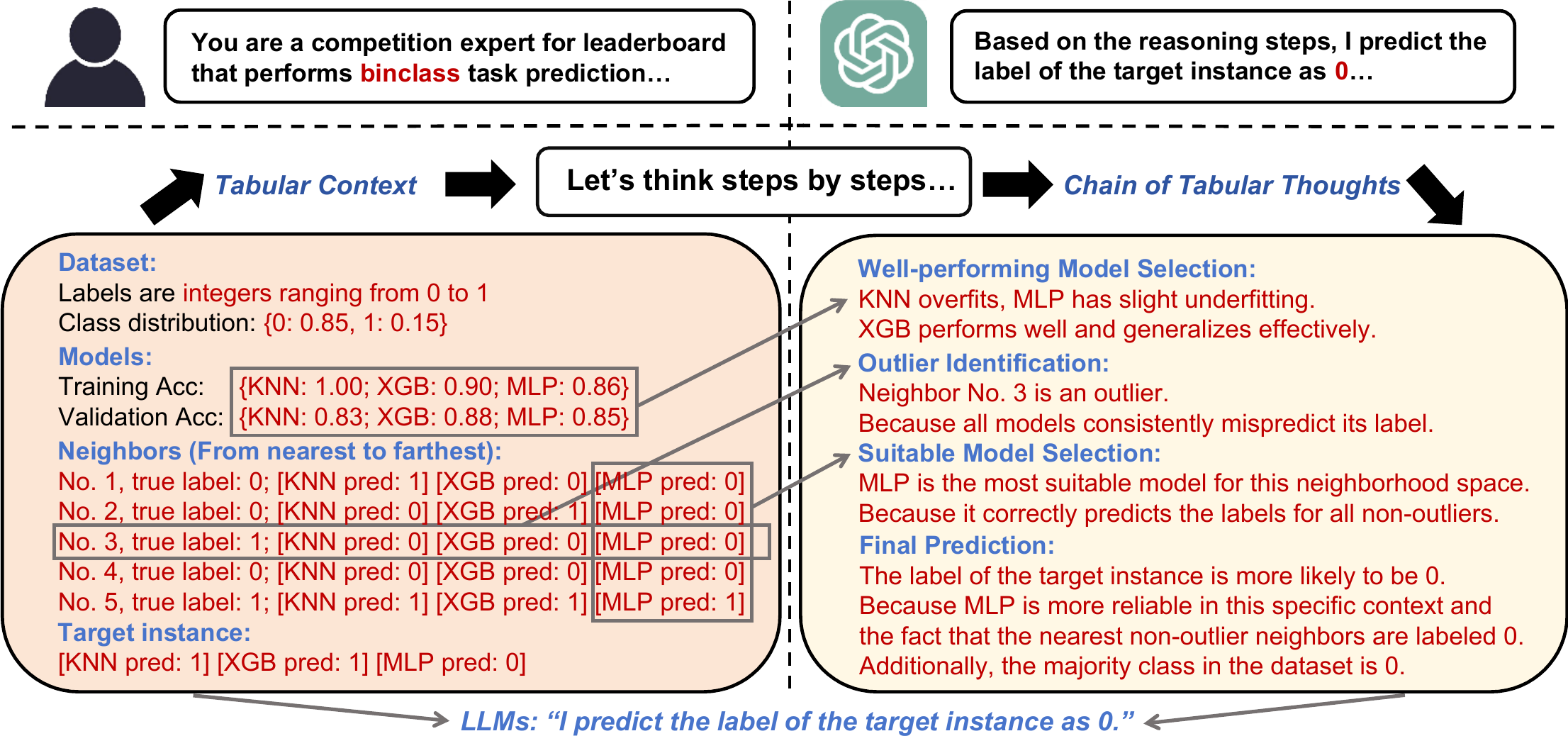}
  \caption{An example of a binary classification task using the tabular context and Chain of Tabular Thoughts (\name). We construct the tabular context based on the combination of neighbors and external model predictions. We design reasoning steps by learning from the thought processes of leaderboard experts. Experts typically first filter models and neighbors, then make predictions by aggregating the external models' predictions for the neighbors and target instances. The tabular context and \name are both provided as a prompt to the LLMs.~\autoref{fig:prompt} shows an example.
  }
  \vspace{-5mm}
\end{figure}

Our goal is to leverage LLMs to perform instance-wise ensemble by reasoning over a structured ``tabular context,'' which conveys alternative forms of knowledge without relying on raw features or semantic descriptions. 
We begin by discussing the background of tabular data learning. 
Then, we introduce how to create a tabular context around the target instance without textual descriptions. Based on this context, we explain the design of the Chain of Tabular Thought (\name), which allows the LLMs to reason clearly.

\subsection{Preliminary}
\noindent{\bf Learning with Tabular Data}. 
Given a labeled tabular dataset $\sD=\{(\x_i, y_i)\}_{i=1}^N$ with $N$ examples (rows in the table). An instance $\x_i$ is associated with a label $y_i$. We consider three types of tasks: binary classification $y_i\in\{0,1\}$, multiclass classification $y_i\in [C]=\{1,\ldots,C\}$, and regression $y_i\in \bbR$. There are $D$ features (columns) for an instance $\x_i$, we denote the $j$-th feature in tabular dataset as $\x_{:,j}$ and denote the $j$-th dimension of $\x_{i}$ as $\x_{ij}$. We learn multiple tabular models $\sM=\{f_m\}_{m=1}^M$ on $\sD$ that each $f_m$ maps $\x_i$ to its label $y_i$. These models exhibit varying generalization capabilities on unseen instances sampled from the same distribution as $\sD$. For example, KNN, XGBoost~\cite{chen2016xgboost}, and Multi-Layer Perceptrons (MLP) are some of the classic models in $\sM$.

\noindent{\bf Predicting with Large Language Models}. 
To make predictions on tabular data using LLMs, we need to generate a prompt $p_i$ containing the necessary information based on the target instance $\x_i$. Existing methods often construct $p_i$ by utilizing feature descriptions $\{F_i\}_{i=1}^D$ and information of dataset $\sD$. For example, in TabLLM~\cite{hegselmann2023tabllm}, $p_i$ includes a textual enumeration of all features. The textual serialization of the $j$-th feature in instance $\x_i$ is ``The feature name $F_j$ is value $x_{ij}$.'' The large language model $\texttt{LLM}$ with vocabulary $\sV$ generates output text $\texttt{LLM}(p_i) \in \sV^*$, which has to be mapped to a valid class in $[C]$ when performing classification. However, when the number of features $D$ is large, the length of the prompt can exceed limitations, and textual descriptions of the dataset may not be available due to data privacy issues or difficulties associated with data collection. To enable the broad application of LLMs in tabular data, we need prompts that do not rely on textual descriptions.

\subsection{Tabular Context Based on Neighbors and External Models}
  To eliminate the limitations imposed by feature descriptions $\{F_i\}_{i=1}^D$ and task descriptions of $\sD$, we need to include substitutes for these textual descriptions in the prompt $p_i$. We use re-weighted distance to search for the target instance's nearest neighbors and initially construct a local context. After that, we incorporate predictions from external models into the local context and add other important information to create the final ``tabular context.''

\noindent{\bf Nearest Neighbor Search}. Due to the non-sequential nature of tabular data, tabular data does not have an inherent context. We address this by finding an implicit sequence based on the distance between instances. We calculate the re-weighted distance between the target instance $\x_i$ to instance $\x_j$ in $\sD$:\looseness=-1
\begin{equation}
    \texttt{dist}(\x_i, \x_j) = \left(\sum_{l=1}^D w_l \cdot |\x_{il} - \x_{jl}|^d\right)^{\frac{1}{d}}\;.
    \label{eq:weighted_dist}
\end{equation}
We set $d=1$ and $w_l > 0$ is a weight for each dimension. When $w_l = 1$, the distance in~\autoref{eq:weighted_dist} degenerates to Manhattan distance ($d=1$). From the labeled dataset $\sD$, we calculate feature weights $w_l$ based on the mutual information~\citep{Brown2012Conditional} between features and labels: $w_l = \texttt{norm}\left(\texttt{mutual}(\x_{:l},\; y)\right)\;$, where $\texttt{norm}\left(\cdot\right)$ normalizes the weights $\{w_l\}_{l=1}^D$ using a min-max scaling method. We rank the distances to obtain the $K$ nearest neighbors $\{\x_1,\x_2,\ldots,\x_K\}$, and their corresponding labels $\{y_1,y_2,\ldots,y_K\}$. The re-weighted distance ensures that neighbors are more similar in important aspects, leading to more meaningful neighbors. The local similarity of neighbors helps provide a relevant and focused context for the target instance. This context can help understand local decision boundaries, leading to more precise and tailored predictions.

\noindent{\bf External Models Integration}. 
 External tabular models can provide additional information and compensate for LLMs' numerical reasoning weaknesses. Therefore, we incorporate external models $\sM=\{f_m\}_{m=1}^M$ on $\sD$ to enrich the context and perform model ensembling. To better apply our method to large datasets, we avoid including feature values in the context, as this would inevitably constrain the prompt length. The knowledge between feature values and labels learned by the external models helps mitigate this information loss. An expert can more accurately infer the most suitable external models for the target instance by analyzing the relationship between the neighbors' true labels and the model predictions. Consequently, we combine the capabilities of trained traditional tabular models with the in-context learning abilities of LLMs. Based on neighbors and external models, the tabular context in our designed prompt $p_i = \texttt{context}\left(\{y_j\}_{j=1}^N,\; \{\left(\x_j,y_j\right)\}_{j=1}^K,\;\sM \right)$ includes:\looseness=-1
\begin{itemize}[noitemsep,topsep=0pt,leftmargin=*]
    \item The basic meta information of dataset $\sD$, such as the label set $[C]$ in classification and the label range in regression. In classification tasks, we include the label frequencies $\{q_i\}_{i=1}^C$ in $\sD$, where $q_i = \sum_{j=1}^N \mathbb{I}(y_j = i)/N $ and $\mathbb{I}(\cdot)$ is the indicator function. 
    \item The training accuracies $\{\texttt{train\_acc}(f_m)\}_{m=1}^M$ and validation accuracies $\{\texttt{val\_acc}(f_m)\}_{m=1}^M$ of each model. These elements are already saved during the construction of $\sM$, and both the training and validation sets come from the partitioning of dataset $\sD$, without introducing additional data.
    \item The true labels of these neighbors $\{y_j\}_{j=1}^K$, and the predictions of $M$ external models for these neighbors $\{\{f_m(\x_j)\}_{m=1}^M\}_{j=1}^K $. These elements can be obtained through $\{\left(\x_j,y_j\right)\}_{j=1}^K$ and $\sM$.
    \item The external models' predictions for target instance $\{f_m(\x_i)\}_{m=1}^M $. 
\end{itemize}

Without including semantic content, we have constructed a tabular context rich in information within the prompt. We anticipate that the robust expert knowledge of LLMs will be able to synthesize this evidence and carry out instance-wise model integration for target instance $x_i$:
\begin{equation}
    \hat{y}_i = \texttt{map}\left( \texttt{LLM}(p_i)\right) = \texttt{map}\left(\texttt{LLM}\left(\texttt{context}\left(\{y_j\}_{j=1}^N,\;\{\left(\x_j,y_j\right)\}_{j=1}^K,\;\sM \right)\right)\right) ,
    \label{eq:tabular_context}
\end{equation}
where $\texttt{map}(\cdot)$ extracts the final prediction from the LLM's response through regular expression matching. For example, in classification tasks, we inform LLMs that we will extract the label from their response using the following code:
\begin{verbatim}
label = re.search(r`I predict the label of the target instance as (\d+)',
                  your_response_text).group(1)
\end{verbatim}

\subsection{Chain of Tabular Thoughts Based on Tabular Context}
\label{sec:cot2}
LLMs often struggle with multi-step or complex reasoning tasks. Our experiments in~\autoref{tab:cot} find that it is challenging for LLMs to directly derive accurate answers from our tabular context. The CoT helps by breaking down the problem into smaller tasks, allowing the model to focus on each step individually. Therefore, we emulate an expert's analysis on the leaderboard and add some reasoning steps to prompt $p_i$. We design the Chain of Tabular Thoughts (\name) to help LLMs reason within our tabular context. Take classification for example, our reasoning steps are as follows:
\begin{enumerate}[label=\alph*),noitemsep,topsep=0pt,leftmargin=*]
\item \noindent{\bf Well-performing Model Selection}. Based on the training accuracies $\{\texttt{train\_acc}(f_m)\}_{m=1}^M$ and validation accuracies $\{\texttt{val\_acc}(f_m)\}_{m=1}^M$ of each model, LLMs infer the overall performance of the external models on the dataset. Then LLMs select $M^w$ well-performing models $\{f_m\}_{m=1}^{M^w}$ from external models:
\begin{equation}
    \{f_m\}_{m=1}^{M^w} = \texttt{step\_a}\left(\{\texttt{train\_acc}(f_m)\}_{m=1}^M,\; \{\texttt{val\_acc}(f_m)\}_{m=1}^M \right).
\end{equation}

We aim for LLMs to identify overfitting and underfitting models based on their training and validation accuracies, and to find the overall well-performing models on $\sD$.
\item \noindent{\bf Outlier Identification}. Based on the true labels of the neighbors $\{y_i\}_{i=1}^K$, the neighbors' predicted labels from well-performing models  $\{\{f_m(\x_j)\}_{m=1}^{M^w}\}_{j=1}^K $, and the label frequencies $\{q_i\}_{i=1}^C$, LLMs identify non-outliers $\{y_j\}_{j=1}^{K^*}$ among the neighbors. Here, we use the true labels to refer to the neighbors:
\begin{equation}
\{y_j\}_{j=1}^{K^*}= \texttt{step\_b}\left(\{y_i\}_{i=1}^K,\;\{\{f_m(\x_j)\}_{m=1}^{M^w}\}_{j=1}^K, \; \{q_i\}_{i=1}^C \right).
\end{equation}
If the majority of well-performing models predict incorrectly for a particular neighbor, it suggests that this neighbor might be an outlier, negatively affecting the predictions. We want the LLMs to be able to identify such outliers. Label frequencies provide additional information about the degree of data imbalance, which aids in reasoning.
\item \noindent{\bf Suitable Model Selection}. Based on the true labels of the non-outliers $\{y_j\}_{j=1}^{K^*}$, the non-outliers' predicted labels from all models $\{\{f_m(\x_j)\}_{m=1}^{M}\}_{j=1}^{K^*}$, and the label frequencies $\{q_i\}_{i=1}^C$, LLMs select $M^s$ the most suitable models $\{f_m\}_{m=1}^{M^s}$ for the neighborhood space of the target instance:
\begin{equation}
\{f_m\}_{m=1}^{M^s} = \texttt{step\_c}\left(   \{y_j\}_{j=1}^{K^*},\; \{\{f_m(\x_j)\}_{m=1}^{M}\}_{j=1}^{K^*} ,\;  \{q_i\}_{i=1}^C      \right).
\end{equation}
Models that perform well overall on the dataset may not be the most efficient at predicting the target instance. It is essential to identify the best-suited models for the target instance within the neighbor space after filtering out outliers.
\item \noindent{\bf Final Prediction}. Based on the true labels of the non-outliers $\{y_j\}_{j=1}^{K^*}$, the label frequencies $\{q_i\}_{i=1}^C$, and the target instance $\x_i$'s predicted labels from the most suitable models and well-performing models $\{f_m(\x_i)\}_{m=1}^{M^s} \cup \{f_m(\x_i)\}_{m=1}^{M^w} $, LLMs make the prediction for the target instance's label $\hat{y}_i$:
\begin{equation}
\hat{y}_i = \texttt{step\_d}\left(  \{y_j\}_{j=1}^{K^*},\; \{f_m(\x_i)\}_{m=1}^{M^s} \cup \{f_m(\x_i)\}_{m=1}^{M^w},\;  \{q_i\}_{i=1}^C      \right).
\end{equation}
After removing outliers and unsuitable external models, LLMs can use a KNN-based approach and model ensembling within the clean local context to achieve the most confident final predictions. Well-performing models, being the strongest models on the current dataset leaderboard, provide auxiliary information for the final prediction.
\end{enumerate}
Finally, we summarize the reasoning steps into text $t$ and include them in prompt $p_i$ in~\autoref{eq:tabular_context}. The tabular context and the Chain of Tabular Thoughts are combined into the final prompt $\Tilde{p}_i$, which is then input into the LLMs to obtain the final prediction:
\begin{equation}
    \hat{y}_i = \texttt{map}\left( \texttt{LLM}(\Tilde{p}_i)\right) = \texttt{map}\left(\texttt{LLM}\left(p_i \cup t \right)\right) .
    \label{eq:cot}
\end{equation}

\noindent{\bf Variant for Regression Tasks}. For the regression task, we remove the label frequency, retain the true labels and model predictions to four decimal places. We use RMSE instead of accuracy. The regular expression for $\texttt{map}(\cdot)$ was changed to $\texttt{(-?\\d+\\.\\d+)}$.~\autoref{fig:prompt_reg} shows an example of the prompts. If the match fails, we will re-enter the prompt until it succeeds. During our experiments, there was no instance of consecutive matching failures occurring 10 times.

\noindent{\bf A Simple Alternative Approach.}  
To assess the necessity of introducing large language models, we design a non-LLM baseline named \textbf{MetaXGB}, which utilizes the same components as our constructed tabular context. For each target instance $\x_i$ in the validation or test set, we retrieve its $K$ nearest neighbors from the training set using the re-weighted distance in~\autoref{eq:weighted_dist}, and collect their true labels as well as the external model predictions on both the target and its neighbors. These components are concatenated into a fixed-length feature vector:
\begin{equation}
\mathbf{z}_i = \left[ \{f_m(\x_i)\}_{m=1}^M, \{y_j\}_{j=1}^K, \{\{f_m(\x_j)\}_{m=1}^M\}_{j=1}^K \right],
\end{equation}
which is then used to train a downstream XGBoost classifier on the validation set. The trained model is evaluated on the test set, and results are compared with our proposed method in~\autoref{sec:results}.

\noindent{\bf Hard Sample Identification.}  
As discussed in~\autoref{sec:intro}, a major challenge in deploying LLM-based tabular methods is their high inference cost, as the LLM must be invoked for each instance. However, in many real-world scenarios, most samples can already be accurately predicted by multiple external models with high agreement. To reduce computational overhead, we adopt a selective strategy that reserves LLM reasoning for more difficult cases—those where external models disagree.
Taking classification tasks as an example, we define a \textit{hard sample} as one for which fewer than a fraction $\tau$ of the $M$ external models predict the same class label. In other words, if more than $\tau$ of the models agree on the prediction, the instance is considered \textit{easy}, and LLM inference can be skipped.

\noindent{\bf Summary}. To address the three challenges of applying LLMs to tabular data, \name introduces the following solutions: 
\begin{itemize}[noitemsep,topsep=0pt,leftmargin=*]
    \item \name designs an information-rich tabular context to replace textual descriptions, freeing LLMs from relying on dataset semantics.
    \item \name helps LLMs leverage the capabilities of external models to understand the numerical relationships between features and labels. Additionally, clear reasoning steps are included to assist LLMs in understanding the relationship between model predictions, neighbor labels, and target predictions.
    \item To reduce inference cost and avoid token limits from including raw features, \name adopts a selective strategy: LLMs are only invoked for hard instances where external models disagree, while easy cases are handled without LLM reasoning.
\end{itemize}

\section{Experiments} \label{sec:Experiments}

\subsection{Setups}
\label{sec: setup}
\noindent{\bf Datasets}. 
To evaluate the effectiveness of \name on challenging tabular prediction tasks, we adopt the TinyBench2 Benchmark Suite~\cite{Ye2024Closer}, a representative subset of 45 datasets selected from a larger benchmark containing over 300 datasets. The full benchmark is designed for evaluating tabular models across diverse data types and task settings. However, due to its scale, it poses a high computational burden for model evaluation.
TinyBench2 addresses this challenge by selecting 15\% of datasets while preserving the relative ranking of models. The selection process is framed as an optimization problem: minimizing the mean absolute error (MAE) between average model ranks on the subset and the full benchmark. The final TinyBench2 shows the best consistency on both seen and unseen models. By using TinyBench2, we efficiently evaluate our method while ensuring the results are representative of full-scale benchmarks~\cite{Ye2024Closer}.

\noindent{\bf \textit{Remark}}.
\name does not require providing dataset descriptions or raw feature values as input to the LLM. Instead, the LLM context is constructed solely from the predictions of external models on the target test sample and the labels and predictions of its nearest neighbors. As a result, we do not need to consider potential dataset leakage during LLM pretraining, nor do we require dedicated dataset leakage detection procedures when selecting evaluation datasets. This makes our approach more broadly applicable, especially when using proprietary or privacy-sensitive tabular data~\cite{Elephants2024,LLMSimpleFeature}.

{\bf Model Set Selection}. To ensure a comprehensive and robust evaluation of ensemble performance, we construct a model set that spans multiple paradigms of tabular modeling. Our goal is twofold: to cover the dominant families of models used in practice, and to expose the ensemble mechanism to diverse inductive biases. Specifically, we include:
\begin{enumerate}[label=\alph*),noitemsep,topsep=0pt,leftmargin=*]
    \item \textbf{Three representative GBDT models}: XGBoost~\cite{chen2016xgboost}, LightGBM~\cite{ke2017lightgbm}, and CatBoost~\cite{Prokhorenkova2018Catboost}, which are widely recognized as state-of-the-art models for tabular data due to their strong performance, robustness, and widespread adoption in both academia and industry.
    
    \item \textbf{Four deep learning models for tabular data}: MLP, ResNet, and FT-Transformer~\cite{GorishniyRKB21Revisiting}, which are representative architectures selected by~\cite{Ye2024Closer} based on systematic benchmarking. To broaden architectural diversity, we also include AutoInt~\cite{SongS0DX0T19AutoInt}, a hybrid model bridging tabular deep learning and recommender systems that integrates attention mechanisms and feature interaction modeling.\looseness=-1
    \item \textbf{A classical non-parametric method}: K-Nearest Neighbors (KNN), which provides an intuitive, instance-based learning paradigm. Including KNN complements the parametric models and offers a contrasting local inductive bias that is useful for diversity in ensemble behavior.
\end{enumerate}
All models are trained independently on each dataset, and their predictions are used by our method and the baselines to construct tabular contexts and evaluate ensemble performance. This carefully chosen model set balances accuracy, architectural diversity, and modeling philosophy.\\
\noindent{\bf Comparison Methods.}  
We compare two main categories of methods, both derived from a common set of tabular models that also serve as the external model pool for \name:
\begin{itemize}[noitemsep,topsep=0pt,leftmargin=*]
    \item \textbf{TinyBench2 Baseline Methods}: This category includes all baseline methods reported in the TinyBench2 benchmark~\cite{Ye2024Closer}, which already cover all the models in our model set. These include classical machine learning models, gradient boosted decision trees (GBDTs), and deep learning architectures for tabular data. In addition, we also compare against TabM~\cite{Yury2024TabM}, a recently proposed deep ensemble learning method that achieves strong performance.
    \item \textbf{Ensemble Methods over the Model Set}: Based on the same model set, we implement several standard ensemble or selection strategies for comparison:
    \begin{itemize}[noitemsep,topsep=0pt,leftmargin=1em]
        \item Best Model: selects the model with the highest validation accuracy on each dataset;
        \item Average Voting: averages the predicted logits across models;
        \item Weighted Voting: averages logits weighted by each model’s training accuracy.
    \end{itemize}
    \item \textbf{Non-LLM Context-based Baseline (MetaXGB)}: We further compare with \textbf{MetaXGB} (see~\autoref{sec:cot2}), a simple non-LLM baseline using the same tabular context with~\name.
\end{itemize}

\noindent{\bf Evaluation Protocol.} 
We follow the evaluation protocol proposed in~\cite{Ye2024Closer} to ensure fair and consistent comparisons across all methods. Specifically, we randomly split each dataset into training, validation, and test sets with a ratio of 64\%: 16\%: 20\%. The validation set is used for model selection and early stopping where applicable. 
All methods, including those in our model set and all comparison baselines, are trained and evaluated on the same data splits. 
To account for randomness, we repeat each experiment five times with different random seeds \{0, 1, 2, 3, 4\} and report the average performance on the test set. For classification tasks, we report average accuracy (Acc), and for regression tasks, we report average Root Mean Squared Error (RMSE).

\subsection{Results}
\label{sec:results}

\begin{wrapfigure}{r}{0.5\textwidth} 
    \centering
    \includegraphics[width=\linewidth]{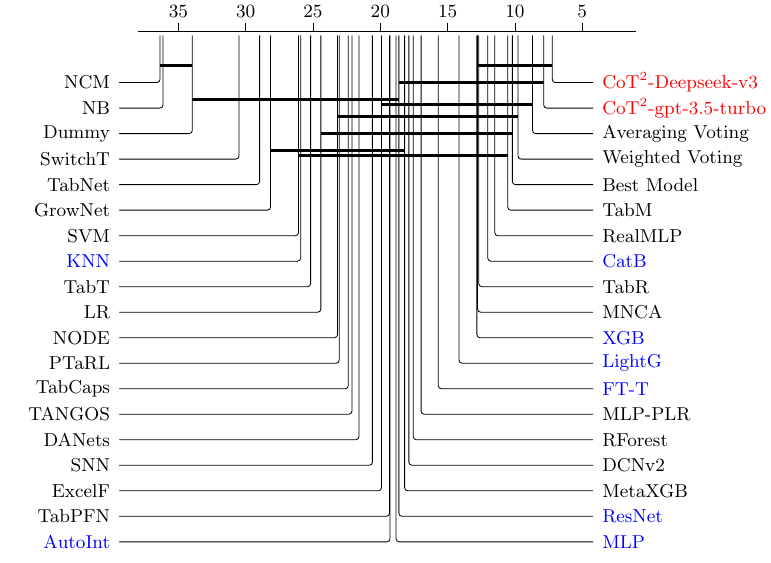}
    \vspace{-5mm}
    \caption{
Critical difference diagram based on the Wilcoxon-Holm test with a significance level of 0.05, used to assess pairwise significance of methods on 30 classification datasets in TinyBench2. 
Blue-colored methods represent the models included in the external model set. The method names in the diagram are abbreviated; the mapping from abbreviations to full names can be found in~\cite{Ye2024Closer} and~\autoref{append:datasets}.
}
    \label{fig:Wilcoxon-Holm}
    \vspace{-5mm}
\end{wrapfigure}

\noindent{\bf Performance on Standard Tasks}.  
For \name, we use \texttt{gpt-3.5-turbo} and \texttt{Deepseek-v3}~\cite{DBLP:journals/corr/abs-2412-19437} with a temperature setting of 0.2. We set the number of neighbors to 10. 
The external models used are shown in~\autoref{sec: setup}. As shown in~\autoref{fig:Wilcoxon-Holm}, Our method achieves the best average ranking across all classification datasets. For regression tasks, the performance results are provided in~\autoref{append:res}.\looseness=-1

\noindent{\bf Effective of \name}. As shown in~\autoref{fig:Wilcoxon-Holm}, \name significantly outperforms the non-LLM baseline MetaXGB, which adopts a hard-rule strategy based on handcrafted feature construction and a downstream XGBoost classifier. This result highlights the limitations of rigid integration methods and demonstrates the necessity of leveraging large language models for more intelligent and flexible model ensembling. 
The key difference between using and not using the chain of tabular thoughts is whether the four inference steps are included in the prompt. 
Incorporating~\name significantly enhances performance when using GPT-3.5 compared to the original tabular context, as shown in~\autoref{tab:cot}. We further include comparisons with \texttt{gpt-4o}, showing that \name continues to bring benefits for more capable models.~\autoref{lab:gpt3.5-turbo} and~\autoref{lab:gpt3.5-turbo-without-cot} show that, without~\name, GPT-3.5's predictions rely solely on the models that perform well on the overall dataset and majority prediction, resulting in an incorrect prediction.~\name enables GPT-3.5 to perform clear and structured reasoning in the tabular context, leading to a correct prediction. 
\begin{table}[t]
\centering
\caption{Mean and STD of test accuracy on five datasets. \name provided significant improvements for GPT-3.5 and smaller benefits for GPT-4o and Deepseek-v3, indicating that the reasoning steps in~\name align well with advanced expert knowledge. ($\textbf{Bold}$ indicates superiority across all methods, while $\underline{\text{underline}}$ signifies whether \name has brought improvements to the same LLM.)}
\resizebox{\textwidth}{!}{
\begin{tabular}{l|cc|cc|cc}
\hline
\multirow{2}{*}{} & \multicolumn{2}{c}{\texttt{gpt-3.5-turbo}} \vline& \multicolumn{2}{c}{\texttt{gpt-4o}} \vline& \multicolumn{2}{c}{\texttt{Deepseek-V3}}\\Dataset
&w/o \name& w/ \name & w/o \name & w/ \name & w/o \name & w/ \name \\ 
\hline
BAS&93.36 $\pm$ 0.28 &  \underline{\textbf{94.63}} $\pm$ 0.18 & 94.40$\pm$ 0.00   & \underline{\textbf{94.63}} $\pm$ 0.18  & 94.10 $\pm$ 0.28 & \underline{94.55} $\pm$ 0.18 \\
DIS&98.41 $\pm$ 0.05 & \underline{98.54} $\pm$ 0.00  & 98.57$\pm$ 0.05 & \underline{98.60} $\pm$ 0.06 & 98.54 $\pm$ 0.15 & \underline{\textbf{98.68}} $\pm$ 0.06 \\
SYL&91.65 $\pm$ 0.18 & \underline{94.56} $\pm$ 0.04 & 94.60$\pm$ 0.30 & \underline{94.87} $\pm$ 0.24
  &   94.43 $\pm$ 0.27 & \underline{\textbf{94.93}} $\pm$ 0.14  \\
CRE&76.37 $\pm$ 0.25 & \underline{77.86} $\pm$ 0.08  & 77.99$\pm$ 0.15& \underline{\textbf{78.01}}$\pm$ 0.15 & 77.80 $\pm$ 0.15 & \underline{77.97} $\pm$ 0.16  \\
FOR&64.18 $\pm$ 0.26 & \underline{68.65} $\pm$ 0.11 & 69.66$\pm$ 0.19 & \underline{\textbf{70.69}}$\pm$ 0.11 & 69.75 $\pm$ 0.27 & \underline{\textbf{70.69}} $\pm$ 0.11  \\
\hline
Mean&84.79 &86.85 & 87.04 &  \textbf{87.36} & 86.92 &\textbf{87.36} \\
\hline
\end{tabular}
}
\label{tab:cot}
\end{table}

The effectiveness of \name helps bridge the performance gap between GPT-3.5 and GPT-4o in this specific reasoning task, demonstrating that our designed reasoning steps align with the more advanced expert knowledge in GPT-4o. With~\name, our simple and efficient prediction context does not require new or complex knowledge. 
The responses of different LLMs to the same prompt are shown in~\autoref{lab:gpt-4o}, and~\autoref{lab:DeepSeek-v3}. We also include responses from the latest version of \texttt{ChatGPT} in~\autoref{lab:step1},~\autoref{lab:step2},~\autoref{lab:step3}, and~\autoref{lab:step4}. We observe that both \texttt{gpt-4o} and \texttt{Deepseek-v3} tend to provide more fine-grained analysis for each piece of information. In particular, \texttt{Deepseek-v3} and the latest ChatGPT often structure their reasoning in a list format, which enhances interpretability and clarity.\looseness=-1

\noindent{\bf Reducing Inference Cost via Selective LLM Usage}.
As discussed in~\autoref{sec:intro}, a key challenge of LLM-based tabular prediction is the high inference cost, as separate prompts must be processed for each instance~\cite{dinh2022lift, hegselmann2023tabllm, tabula8b}. 
To reduce inference cost, we adopt a strategy to identify easy instances—those for which external models show high agreement. Specifically, for classification tasks, we define an instance as easy if at least \( \tau = 3/4 \) of the external models agree on the prediction, and LLM inference is skipped in these cases.
This selective strategy significantly reduces computational overhead by reserving LLM inference for more challenging instances. As shown in~\autoref{tab:dataset_info}, this approach allows us to bypass LLM reasoning for the majority of test samples, ensuring that LLMs are used only when their reasoning capabilities are most needed.

\subsection{Ablation Study}
To better understand the design choices in \name, we conduct an ablation study on several key components. All the ablation experiments are conducted using the \texttt{gpt-3.5-turbo} model on five classification datasets in TinyBench2, and the results are reported in~\autoref{append:ablation}.

\begin{figure}
    \begin{minipage}[t]{\textwidth}
    \centering
    \includegraphics[width=\textwidth]{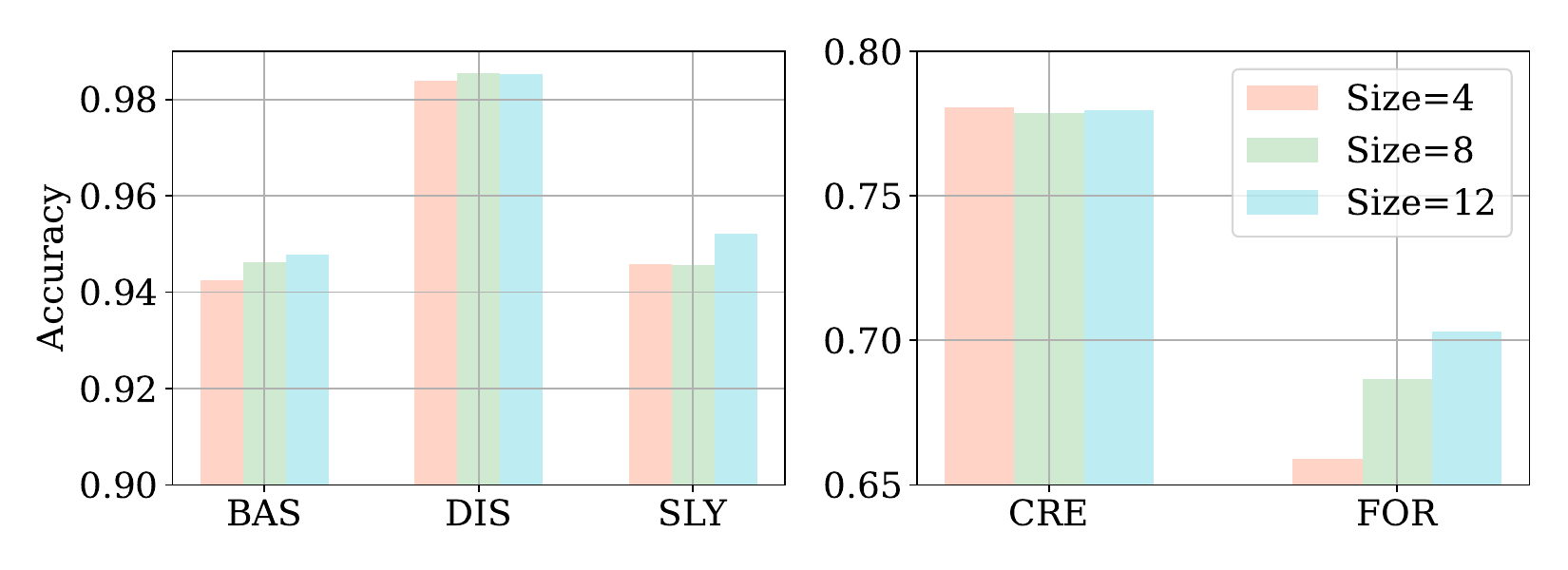}
    \caption{
    Impact of external model set size and quality on the performance of~\name. 
    }
    \label{fig:abl_model_set}    
    \end{minipage}
    \hfill
    \begin{minipage}[t]{\textwidth}
    \centering
    \includegraphics[width=\textwidth]{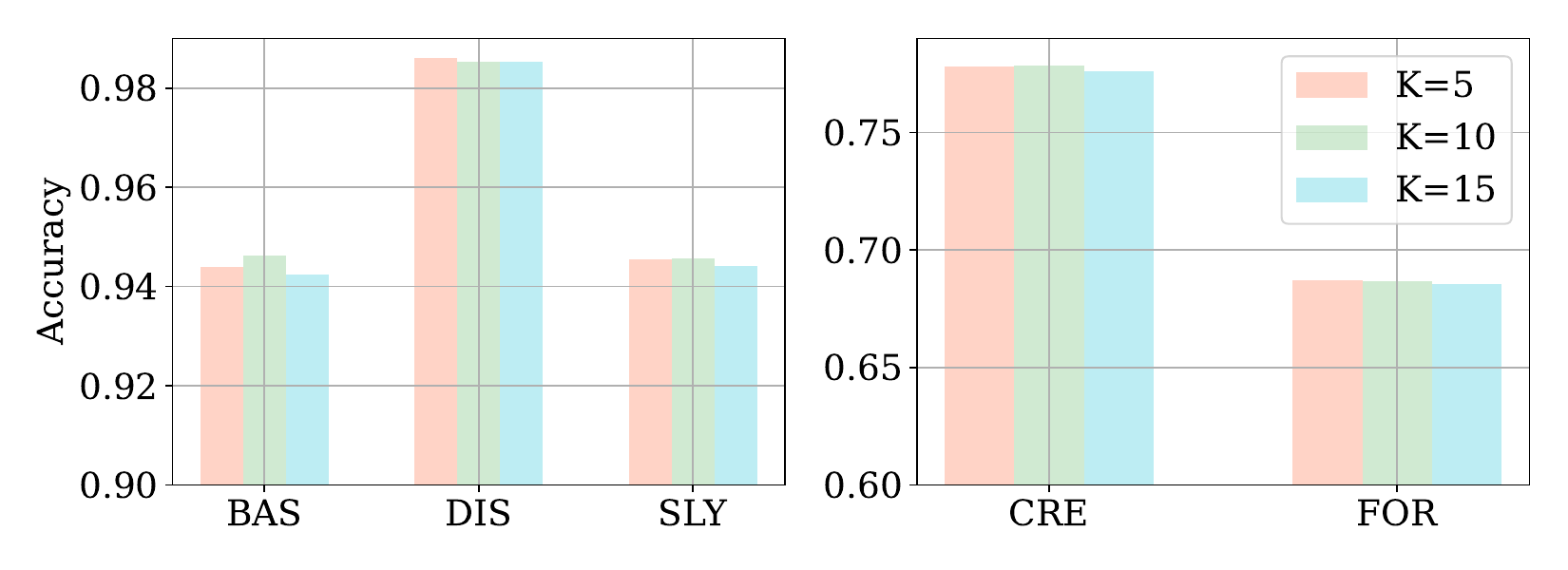}
    \caption{
    Performance of \name\ under different numbers of neighbors $k$ used in the context. 
    }
    \label{fig:abl_k} 
    \end{minipage}
\end{figure}

\textbf{Model Set}: We investigate the impact of the external model set on the performance of~\name by varying both the number and quality of models included. Specifically, we experiment with different pool sizes and progressively introduce stronger models into the ensemble. We evaluate three configurations: a reduced model set of 4 strong models (XGBoost, CatBoost, MLP, FT-Transformer), the original 8-model pool used in the main experiments, and an extended 12-model set that adds four recent, higher-performing deep models (RealMLP~\cite{David2024RealMLP}, TabR~\cite{gorishniy2023tabr}, ModernNCA~\cite{Ye2024ModernNCA}, and TabM~\cite{Yury2024TabM}).  The results demonstrate that increasing the number of models generally enhances performance, while incorporating higher-performing models into the pool leads to further gains, as shown in~\autoref{fig:abl_model_set}. 

\textbf{Number of Neighbors ($k$)}: We evaluate different values of $k$ when constructing the tabular context. The results show that moderate values (\textit{e.g.}, $k=10$) strike a good balance between context richness and prompt length, as shown in~\autoref{fig:abl_k}.

Additionally, we summarize key design choices and ablation factors affecting the performance of~\name, including distance metrics, model name anonymization, LLM inference temperature, and the hard-sample selection threshold. Detailed results and discussions are provided in~\autoref{append:ablation}.

\section{Conclusion} \label{sec: Colclusion}

The widespread use of LLMs on tabular data is limited by several factors: a heavy reliance on textual descriptions, an inability to handle datasets with a large number of features, and insensitivity to numerical values. To apply the expert knowledge of LLMs to aid in predictions on tabular data, we designed a tabular context incorporating instance-specific insights as a substitute for semantic descriptions and feature values. By utilizing the capabilities of external models, we addressed the weaknesses of LLMs in handling the relationship between numerical features and labels. Additionally, we devised a chain of tabular thoughts to teach LLMs how to comprehend numerical values within our tabular context. Our method can be efficiently applied to tabular prediction tasks.

{
\bibliography{main}
\bibliographystyle{plain}
}

\newpage
\appendix
The Appendix consists of four sections:
\begin{itemize}[noitemsep,topsep=0pt,leftmargin=20pt]
\item \autoref{append:datasets}: We provide detailed descriptions of the datasets used in our experiments, along with implementation details for reproducibility.
\item \autoref{append:ablation}: We present a comprehensive ablation study analyzing the impact of key design choices in our method.
\item \autoref{append:res}: We include complete experimental results that were omitted from the main paper due to space limitations.
\item \autoref{append:example}: We show representative examples of our method, including prompt formats and responses under different settings.
\end{itemize}

\section{Datasets and implementation details}
\label{append:datasets}

To facilitate comprehensive evaluation and analysis, we adopt the TinyBench2 benchmark~\cite{Ye2024Closer}, which includes a diverse collection of 45 tabular datasets spanning different task types, including binary classification, multiclass classification, and regression. These datasets vary widely in size, feature composition, and difficulty, making them suitable for robust and fair assessment of tabular learning methods.

\autoref{tab:dataset_info} summarizes the key statistics for each dataset used in our experiments. Specifically, we report the following information:
\begin{itemize}[noitemsep,topsep=0pt,leftmargin=20pt]
    \item \textbf{Abbr}: A short identifier used throughout the paper for concise reference.
    \item \textbf{Task\_type}: The type of machine learning task (regression, binclass, or multiclass).
    \item \textbf{N / C}: The number of numerical and categorical features, respectively.
    \item \textbf{Samples}: The total number of instances in the dataset.
    \item \textbf{Hard ratio}: The percentage of hard samples, indicating the dataset's learning difficulty.
\end{itemize}
We quantify dataset difficulty using the \textbf{hard ratio}, which represents the proportion of hard samples in each dataset. A sample is considered \textit{hard} if it fails to reach consensus among external models during evaluation. For classification tasks, a sample is labeled as hard if fewer than 3/4 of the external models predict the same class label. For regression tasks, we use an outlier-based rule: a sample is marked as hard if more than 1/4 of the external model predictions fall outside the interquartile range (IQR), specifically beyond $[Q_1 - 1.5 \times \text{IQR}, Q_3 + 1.5 \times \text{IQR}]$. These criteria help identify instances that are difficult to predict consistently, providing a measure of dataset complexity.

\begin{table}
    \centering
    \caption{The list of datasets in TinyBench2~\cite{Ye2024Closer},
along with the statistics for each dataset.}
    \label{tab:dataset_info}
    \resizebox{\textwidth}{!}{
\begin{tabular}{lllllll}
\toprule
Dataset & Abbr &   Task\_type &    N &   C & Samples & Hard ratio \\
\midrule
Ailerons                               &  AIL &  regression &   40 &   0 &   13750 &    49.3818 \\
BNG(breast-w)                          &  BWR &    binclass &    9 &   0 &   39366 &     0.5588 \\
BNG(cmc)                               &  CMC &  multiclass &    2 &   7 &   55296 &     9.5931 \\
BNG(tic-tac-toe)                       &  TTT &    binclass &    0 &   9 &   39366 &     4.6355 \\
CPMP-2015-regression                   &  C2R &  regression &   23 &   2 &    2108 &    53.7915 \\
Cardiovascular-Disease-dataset         &  CDD &    binclass &    5 &   6 &   70000 &     3.9214 \\
CookbookReviews                        &  COO &  regression &    7 &   0 &   18182 &     4.1793 \\
FOREX\_audchf-day-High                  &  ADH &    binclass &   10 &   0 &    1833 &    28.0654 \\
FOREX\_audsgd-hour-High                 &  AHH &    binclass &   10 &   0 &   43825 &    26.5830 \\
FOREX\_cadjpy-hour-High                 &  FOR &    binclass &   10 &   0 &   43825 &    21.5402 \\
Gender\_Gap\_in\_Spanish\_WP               &  GGI &  multiclass &   13 &   0 &    4746 &    10.5263 \\
IEEE80211aa-GATS                       &  IGE &  regression &   27 &   0 &    4046 &    46.7901 \\
KDD                                    &  KDD &    binclass &   34 &  11 &    5032 &    12.1152 \\
Large-scale\_Wave\_Energy\_Farm\_Sydney\_49 &  LSW &  regression &   99 &   0 &   17964 &    43.6961 \\
Superconductivty                       &  SUP &  regression &   81 &   0 &   21197 &    37.9953 \\
VulNoneVul                             &  VUL &    binclass &   16 &   0 &    5692 &     0.0000 \\
archive2                               &  ARC &  regression &   11 &   1 &    1143 &    34.0611 \\
bank8FM                                &  BAN &  regression &    8 &   0 &    8192 &    64.2465 \\
baseball                               &  BAS &  multiclass &   15 &   1 &    1340 &     1.4925 \\
communities\_and\_crime                  &  CAC &  regression &  102 &   0 &    1994 &    36.3409 \\
credit                                 &  CRE &    binclass &   10 &   0 &   16714 &    10.5594 \\
dis                                    &  DIS &    binclass &    6 &  23 &    3772 &     0.3974 \\
eye\_movements\_bin                      &  EMB &    binclass &   20 &   0 &    7608 &    25.9527 \\
fried                                  &  FRI &  regression &   10 &   0 &   40768 &    71.4251 \\
healthcare\_insurance\_expenses          &  HIE &  regression &    3 &   3 &    1338 &    27.9851 \\
house\_16H\_reg                          &  H1R &  regression &   16 &   0 &   22784 &    32.1922 \\
jungle\_chess\_2pcs\_raw\_endgame\_complete &  JC2 &  multiclass &    6 &   0 &   44819 &    12.0259 \\
kin8nm                                 &  KIN &  regression &    8 &   0 &    8192 &    33.3130 \\
law-school-admission-bianry            &  LSA &    binclass &    7 &   4 &   20800 &     0.0000 \\
mfeat-fourier                          &  MFF &  multiclass &   76 &   0 &    2000 &    11.0000 \\
mv                                     &   MV &  regression &    7 &   3 &   40768 &    91.2558 \\
online\_shoppers                        &  OSN &    binclass &    5 &   9 &   12330 &     4.9067 \\
page-blocks                            &  PBA &  multiclass &   10 &   0 &    5473 &     1.4612 \\
pc3                                    &  PC3 &    binclass &   37 &   0 &    1563 &     3.8339 \\
pendigits                              &  PEN &  multiclass &   16 &   0 &   10992 &     0.5457 \\
qsar\_fish\_toxicity                     &  QFT &  regression &    4 &   2 &     908 &    31.3187 \\
rl                                     &   RL &    binclass &    5 &   7 &    4970 &    24.3461 \\
satimage                               &  SAT &  multiclass &   36 &   0 &    6430 &     5.5210 \\
segment                                &  SEG &  multiclass &   17 &   0 &    2310 &     6.2771 \\
sylvine                                &  SYL &    binclass &   20 &   0 &    5124 &     3.3171 \\
taiwanese\_bankruptcy\_prediction        &  TBP &    binclass &   95 &   0 &    6819 &     0.8798 \\
waveform-5000                          &  W5A &  multiclass &   40 &   0 &    5000 &     6.7000 \\
website\_phishing                       &  WPE &  multiclass &    0 &   9 &    1353 &     8.1181 \\
wine-quality-white                     &  WQW &  multiclass &   11 &   0 &    4898 &    24.5918 \\
yeast                                  &  YEA &  multiclass &    8 &   0 &    1484 &    16.1616 \\
\bottomrule
\end{tabular}
}
\end{table}

\noindent{\bf External models.}  
For all external baseline models that do not explicitly specify preprocessing strategies for categorical and numerical features—such as MLP and ResNet—we uniformly apply one-hot encoding for categorical features and standard normalization for numerical features and regression labels. Training is performed with a maximum of 200 epochs, a batch size of 1024, and early stopping with a patience of 20 epochs.  
We conduct 100 rounds of hyperparameter tuning for each external model~\cite{Liu2024Talent}. The full search space configurations are available at \url{https://github.com/LAMDA-Tabular/TALENT/tree/main/TALENT/configs/opt_space}.

\noindent{\bf \name Configuration.}  
In our main experiments, we run each dataset using 5 different random seeds and report the average accuracy (for classification) or RMSE (for regression). For each target instance, we retrieve $k=10$ nearest neighbors from the training set as context, and set the temperature parameter to $t=0.2$. We deploy the \name pipeline using two large language models: \texttt{gpt-3.5-turbo-0125} and \texttt{DeepSeek-V3-P001}. Additionally, results from \texttt{gpt-4o} are reported in~\autoref{tab:cot} for further comparison.

{\bf Abbreviations of models compared in our main experiments.}
We group all baseline methods into several categories for clarity. Classical methods include Dummy, Logistic Regression (LR), K-Nearest Neighbors (KNN), Support Vector Machines (SVM), Naive Bayes, Linear Regression (LR), and DNNR. Tree-based methods include Random Forest (RF), XGBoost (XGB)~\cite{chen2016xgboost}, LightGBM (LightG)~\cite{ke2017lightgbm}, and CatBoost (CatB)~\cite{Prokhorenkova2018Catboost}. MLP variants cover vanilla MLP, MLP-PLR~\cite{Gorishniy2022On}, Self-Normalizing Neural Networks (SNN)~\cite{KlambauerUMH17SNN}, ResNet~\cite{GorishniyRKB21Revisiting}, RealMLP~\cite{David2024RealMLP}, and TabM~\cite{Yury2024TabM}. Special architectures include DCNv2~\cite{WangSCJLHC21DCNv2}, DANets~\cite{ChenLWCW22DAN}, and TabCaps~\cite{Chen2023TabCaps}. Token-based methods include AutoInt~\cite{SongS0DX0T19AutoInt}, TabTransformer (TabT)~\cite{Huang2020TabTransformer}, FT-Transformer (FT-T)~\cite{GorishniyRKB21Revisiting}, and ExcelFormer (ExcelF)~\cite{Chen2023Excel}. Regularization-based methods comprise TANGOS~\cite{jeffares2023tangos}, SwitchTab (SwitchT)~\cite{Wu2024SwitchTab}, and PTaRL~\cite{ye2023ptarl}. Tree-mimic methods include NODE~\cite{PopovMB20Neural}, GrowNet~\cite{Badirli2020GrowNet}, and TabNet~\cite{ArikP21TabNet}. Context-based methods include TabR~\cite{gorishniy2023tabr}, TabPFN~\cite{Hollmann2022TabPFN} and ModernNCA (MNCA)~\cite{Ye2024ModernNCA}.

\section{Ablation Study}
\label{append:ablation}
\begin{figure}
    \centering
    \includegraphics[width=\textwidth]{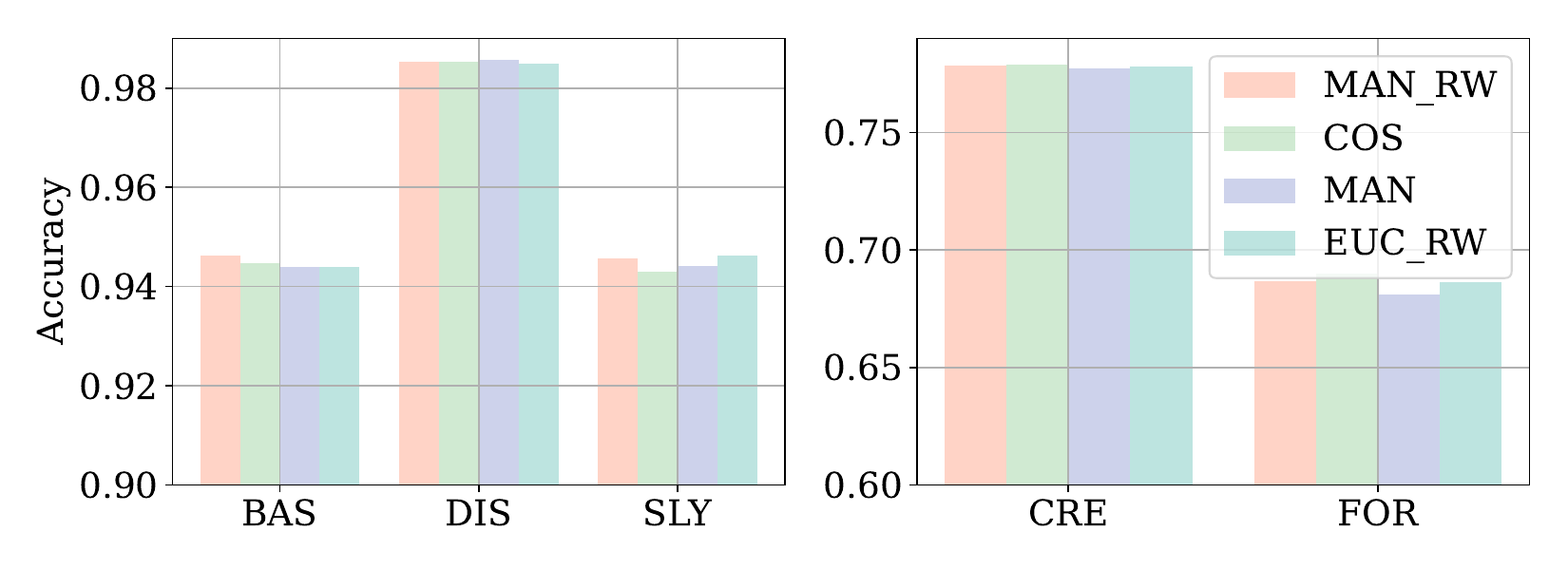}
    \caption{In the process of nearest neighbor search, we used the Manhattan distance reweighted by mutual information (MAN-RW) in the main experiment. We also experimented with cosine distance (COS) and Euclidean distance reweighted by mutual information (EUC-RW). The knowledge from LLMs and the predictions from external models can help us filter out outliers in the nearest neighbors, making~\name robust to different distance metrics.}
    \label{fig:abl_distance}
\end{figure}

\begin{figure}
    \centering
    \includegraphics[width=\textwidth]{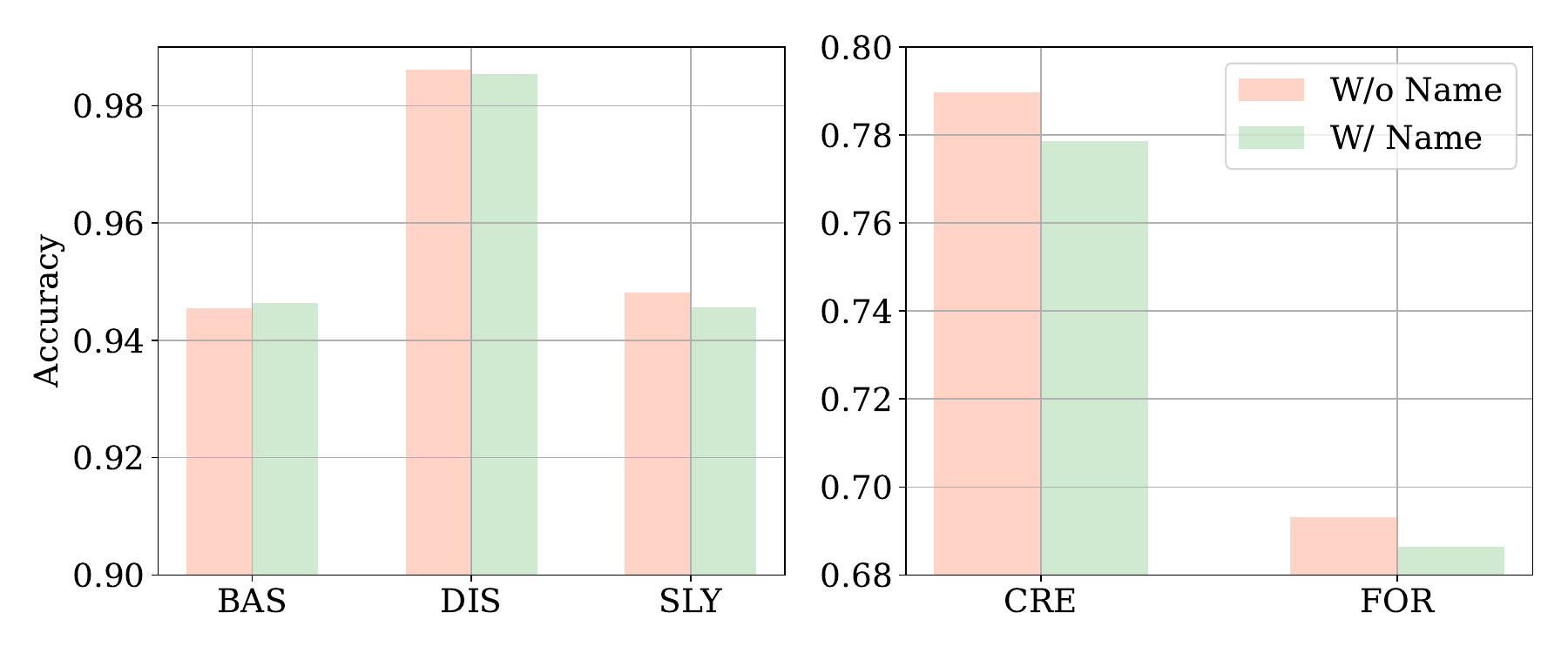}
    \caption{
Impact of anonymizing external model names in the tabular context on \name's performance. 
We compare two settings: \textit{w/ name}, where real model names are provided, and \textit{w/o name}, where anonymized labels (\textit{e.g.}, Model A, B, C) are used. 
Results show that the anonymized version (\textit{w/o name}) outperforms the named version on four out of five datasets, indicating that removing model identity may reduce bias and improve reasoning consistency.
}

    \label{fig:abl_model_name}
\end{figure}

\begin{figure}
    \centering
    \includegraphics[width=\textwidth]{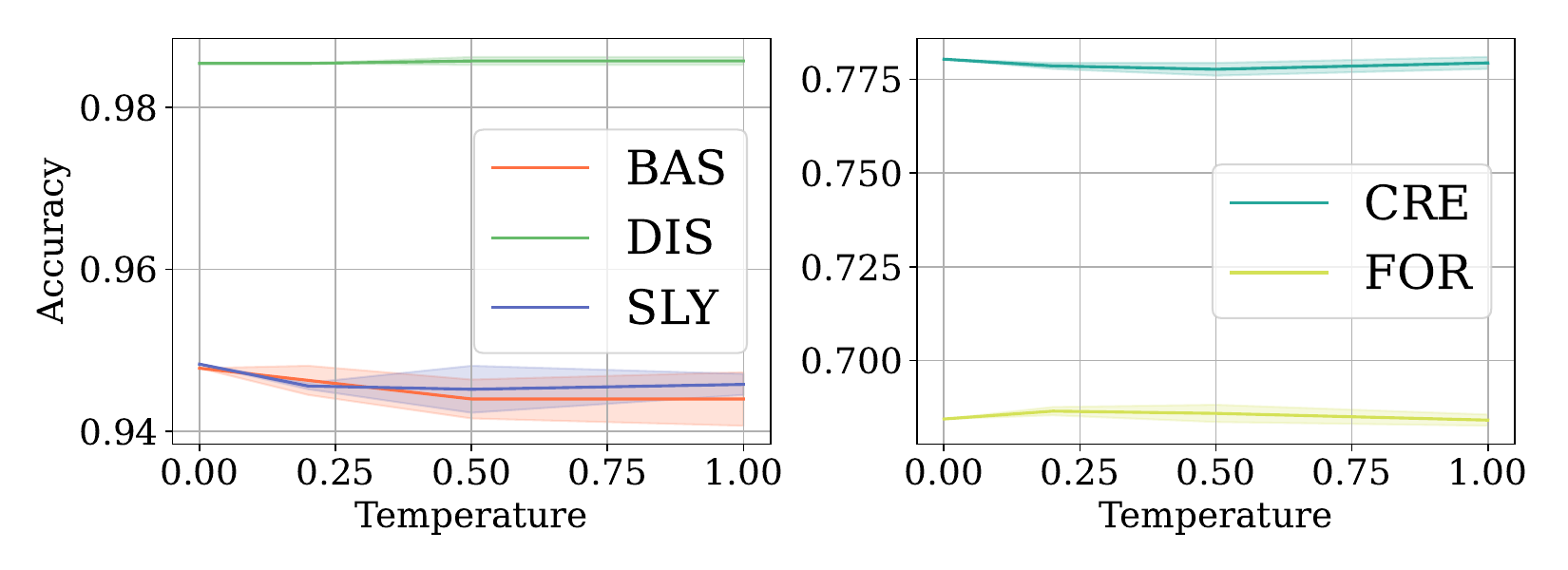}
    \caption{
Effect of temperature setting on \name's performance. We evaluate four values: $t=0.1$, $0.2$, $0.5$, and $1.0$. 
Results show that \name is generally robust to temperature changes, with performance remaining stable across different $t$ values. 
However, higher temperatures lead to increased variance, indicating less stable behavior from the LLM during inference.
}
    \label{fig:abl_temperature}
\end{figure}

The additonal ablation experiments, as shown in~\autoref{fig:abl_distance},~\autoref{fig:abl_model_name},~\autoref{fig:abl_temperature}, and~\autoref{tab:threshold}, are conducted on a subset of five datasets: \textbf{BAS}, \textbf{DIS}, \textbf{SYL}, \textbf{CRE}, and \textbf{FOR}.

\begin{itemize}[noitemsep,topsep=0pt,leftmargin=*]
\item \textbf{Distance Metric}: We compare several distance metrics for neighbor retrieval, including Manhattan, Euclidean, cosine similarity, and our proposed re-weighted distance in~\autoref{eq:weighted_dist}. These metrics affect how relevant neighbors are selected for each target instance, which in turn influences the quality of the constructed tabular context (\autoref{fig:abl_distance}).
\item \textbf{Anonymizing External Model Names}: We examine whether hiding the real names of external models in the tabular context affects \name's performance. Instead of using actual model names, we substitute them with anonymized labels (\textit{e.g.}, Model A, B, C, D). Interestingly, we observe improved performance on four out of five datasets under this anonymized setting. This suggests that LLMs may carry inherent biases or preferences toward certain model names, and removing these cues can lead to more objective and consistent reasoning (\autoref{fig:abl_model_name}).
\item \textbf{LLM Inference Temperature}: We analyze the effect of temperature settings on model outputs. Lower temperatures (\textit{e.g.}, 0.2) yield more stable and deterministic predictions, while higher temperatures introduce variability and may reduce accuracy (\autoref{fig:abl_temperature}).
\item \textbf{Threshold for Hard Sample Selection}: We study how varying the agreement threshold for identifying hard samples affects both predictive performance and inference cost (\autoref{tab:threshold}).
\end{itemize}

\begin{table}[H]
\caption{
Effect of varying the hard sample threshold on accuracy and inference cost. 
Increasing the threshold allows more instances to be handled by the LLM, but may decrease accuracy due to potential hallucinations on simple instances. 
Conversely, decreasing the threshold may also reduce accuracy, as difficult samples not solvable by base ensembles alone may be excluded from LLM inference. 
Results are reported on the SYL and CRE datasets.
}

\label{tab:threshold}
\centering
\begin{tabular}{lcccccc}
\toprule
Dataset & Threshold & Accuracy (\%) & Time (s) & Tokens (input) & Tokens (output) & Price (\$)\\
\midrule
\multirow{3}{*}{SYL} 
& 0.50 & 94.20 & 19.3 & 41715 & 6148 & 0.03 \\
& 0.75 & 94.56 & 62.4 & 74920 & 11841 &0.06\\
& 1.00 & 93.06 & 213.2 & 394769 & 60807& 0.29\\
\midrule
\multirow{3}{*}{CRE} 
& 0.50 & 77.08 & 146.3 & 275757 & 41475 &0.20\\
& 0.75 & 77.86 & 644.8 & 1525206 & 134162&0.96 \\
& 1.00 &  77.29 &1418.4&2995927 &454875 & 2.18\\
\bottomrule
\end{tabular}
\end{table}

\section{Detailed Results}
\label{append:res}

\begin{table}[H]
    \caption{The detailed results shown in~\autoref{fig:Wilcoxon-Holm}.}
    \label{tab:res}
    \centering
    \resizebox{\textwidth}{!}{
    \begin{tabular}{lccc|lccc}
\toprule
Dataset & \name-Deepseek-v3 & MetaXGB & \name-gpt-3.5 & Dataset & \name-Deepseek-v3 & MetaXGB & \name-gpt-3.5 \\
\midrule
BAS& 94.55 & 95.52 & 94.63 &WPE & 92.03 & 88.93 & 91.14 \\
PC3 & 89.14 & 89.14 & 89.39 &
ADH & 74.22 & 71.12 & 69.65 \\
MFF & 87.50 & 86.25 & 88.25 &
SEG & 93.29 & 92.64 & 93.81 \\
DIS & 98.68 & 98.15 & 98.54 &
GGI & 60.40 & 56.21 & 59.64 \\
WQW & 63.59 & 63.67 & 63.84 &
RL & 78.81 & 77.67 & 77.87 \\
W5A & 85.80 & 83.30 & 86.12 &
KDD & 81.15 & 78.35 & 80.10 \\
SYL & 94.93 & 94.05 & 94.56 &
PBA & 97.44 & 96.53 & 97.50 \\
VUL & 98.95 & 98.95 & 98.95 &
SAT & 92.40 & 90.75 & 92.40 \\
TBP & 97.20 & 96.41 & 97.27 &
EMB & 62.67 & 59.86 & 62.65 \\
PEN & 99.43 & 99.18 & 99.45 &
OSN & 90.30 & 89.94 & 90.18 \\
CRE & 77.97 & 74.69 & 77.86 &
LSA & 100.0 & 100.0 & 100.0 \\
BWR & 98.74 & 98.63 & 98.70 &
TTT & 81.47 & 78.54 & 81.52 \\
FOR & 70.69 & 66.53 & 68.65 &
AHH & 68.55 & 66.34 & 65.65 \\
JC2 & 95.26 & 98.57 & 90.13 &
CMC & 58.88 & 55.48 & 58.84 \\
CDD & 73.48 & 70.84 & 73.46 &
YEA & 60.88 & 58.22& 60.94   \\
\bottomrule
\end{tabular}
}
\end{table}

\begin{table}[H]
    \centering
    \caption{RMSE on 15 regression datasets in TinyBench2. We report the RMSE of all the external models for each dataset. \name achieved the highest average ranking among all methods. }
\resizebox{\textwidth}{!}{
\begin{tabular}{lcccccccccc}
\toprule
Dataset & KNN & XGBoost & Catboost & LightGBM & MLP & ResNet & AutoInt & FT-T & Average & \name \\
\midrule
ARC$_{\times 10^2}$ & 3.6422 & 3.3812 & 3.2327 & 3.4980 & 3.6477 & 3.5902 & 3.7367 & 4.0321 & 3.2382 & 3.2491 \\
HIE$_{\times 10^3}$ & 5.5246 & 4.6865 & 4.5222 & 4.6913 & 4.8525 & 4.7755 & 4.8049 & 4.5223 & 4.5460 & 4.6150 \\
CAC$_{\times 10^{-1}}$ & 1.3446 & 1.3502 & 1.2977 & 1.3308 & 1.3584 & 1.4602 & 1.3649 & 1.3791 & 1.3033 & 1.2989 \\
IGE$_{\times 10^{-2}}$ & 8.4527 & 4.2323 & 3.6572 & 4.2587 & 3.0002 & 2.4307 & 2.8165 & 3.0886 & 2.9917 & 2.7597 \\
KIN$_{\times 10^{-1}}$ & 1.2049 & 1.249 & 0.9029 & 1.2599 & 0.7488 & 7.3773 & 7.0919 & 0.6754 & 0.7835 & 0.7699 \\
BAN$_{\times 10^{-2}}$ & 4.9246 & 3.0842 & 2.8628 & 3.0073 &2.8947 & 2.8571 & 2.8360 & 2.8245 & 2.8505 & 2.8142 \\
AIL$_{\times 10^{-4}}$ & 2.0400 & 1.5300 & 1.4700 & 1.5200 & 1.5500 & 1.5500 & 1.5500 & 1.5700 & 1.4800 & 1.4700 \\
LSW$_{\times 10^{4}}$ & 1.1759 & 0.5011 & 0.4449 & 0.4991 & 0.4841 & 0.5963 & 0.6354 & 0.4007 & 0.4249 & 0.3964 \\
COO$_{\times 10^{0}}$ & 1.4921 & 1.4795 & 1.4877 & 1.4833 & 1.5112 & 1.5921 & 1.5777 & 1.5899 & 1.4947 & 1.4951 \\
SUP$_{\times 10^{1}}$ & 1.0713 & 0.9959 & 0.9980 & 1.0103 & 1.0738& 1.0365 & 1.0924& 1.0593 & 0.9744 & 0.9754 \\
H1R$_{\times 10^{4}}$ & 3.7025 & 3.1061 & 3.0191 & 3.1017 & 3.1441 & 3.1448 & 3.1296 & 3.1265 & 2.9075 & 2.9143 \\
MV$_{\times 10^{-1}}$ & 15.1106 & 0.9397 & 0.8157 & 0.9257 & 0.2590 & 1.2554 & 0.4128 & 0.2684 & 1.9614 & 0.4875 \\
FRI$_{\times 10^{0}}$ & 1.8540 & 1.0838 & 1.0105 & 1.0627 & 1.0840 & 1.0230 & 1.0201 & 1.0100 & 1.0330 & 1.0122 \\
QFT$_{\times 10^{-1}}$ & 9.7412 & 9.2242& 8.7159 & 8.9754 & 9.1436 & 9.4800 & 9.1662 & 9.1038 & 8.6904 & 8.9618 \\
C2R$_{\times 10^{2}}$ & 5.7960 & 4.6258 & 4.7438 & 4.7587 & 5.3895 & 5.1601 & 5.3207 & 5.3627 & 4.9601 & 4.9860 \\
\bottomrule
average rank & 8.80 & 5.73 & 3.27& 5.53 & 6.80 & 6.67 & 6.27& 5.27 & 3.67 & 3.00 \\
\bottomrule
\end{tabular}
}
 \label{tab:regression}
    \label{tab:my_label}
\end{table}

\section{Examples}
\label{append:example}
\begin{figure}[H]
    \centering
    \includegraphics[width=0.9\textwidth]{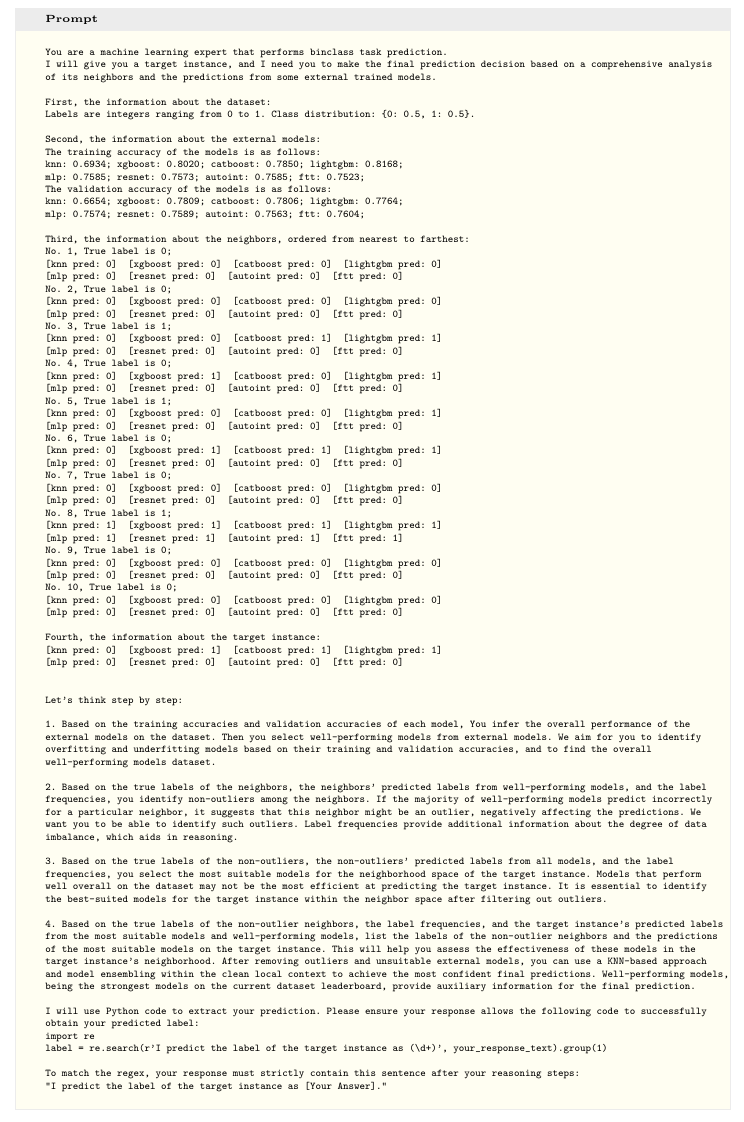}
    \caption{An example of the prompt in the classification dataset CRE. We also provided examples
in~\autoref{lab:gpt3.5-turbo-without-cot} and~\autoref{lab:gpt3.5-turbo} where gpt-3.5-turbo responds without and with~\name, respectively. It
can be observed that~\name breaks down a complex problem into multiple steps, resulting in more
structured answers, thus enhancing the interpretability and accuracy. The responses of Deepseek-v3 and
GPT-4o are in~\autoref{lab:DeepSeek-v3} and~\autoref{lab:gpt-4o}. We further provide step-by-step responses from the latest ChatGPT to illustrate the reasoning process in more detail, as shown in~\autoref{lab:step1},~\ref{lab:step2},~\ref{lab:step3}, and~\ref{lab:step4}.}
    \label{fig:prompt}
\end{figure}

\begin{figure}[H]
    \centering
    \includegraphics[width=\textwidth]{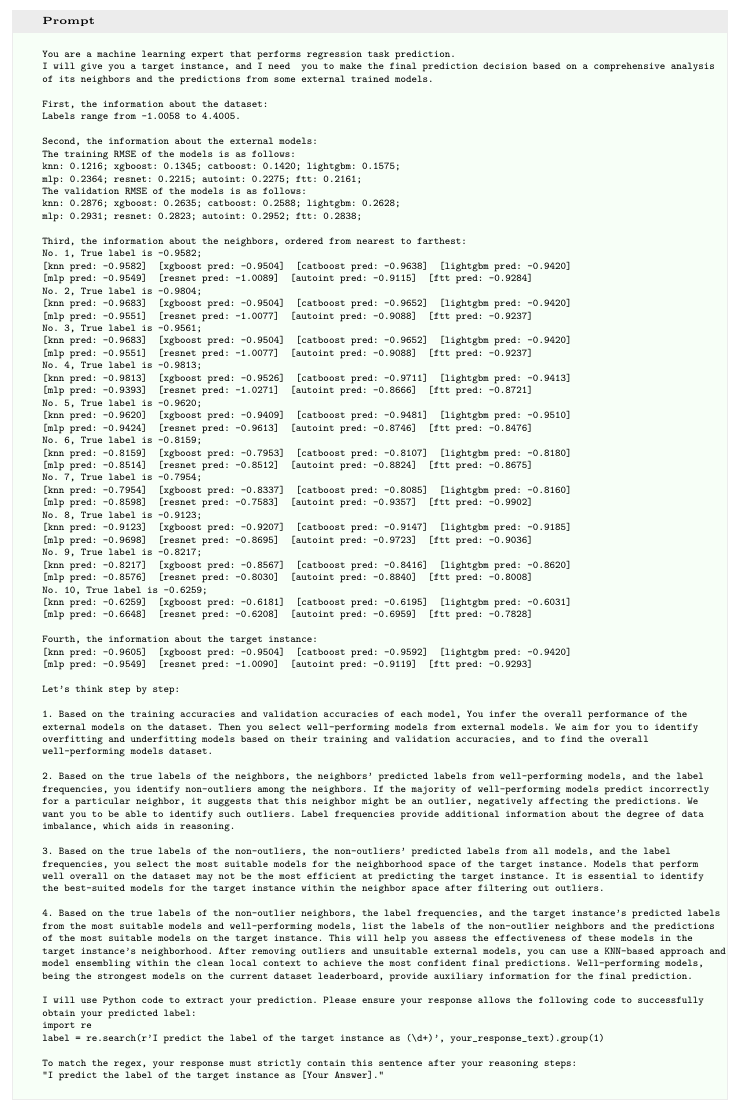}
    \caption{An example of the prompt in the regression dataset KIN.}
    \label{fig:prompt_reg}
\end{figure}

\begin{figure}[h]
    \centering
    \includegraphics[width=\textwidth]{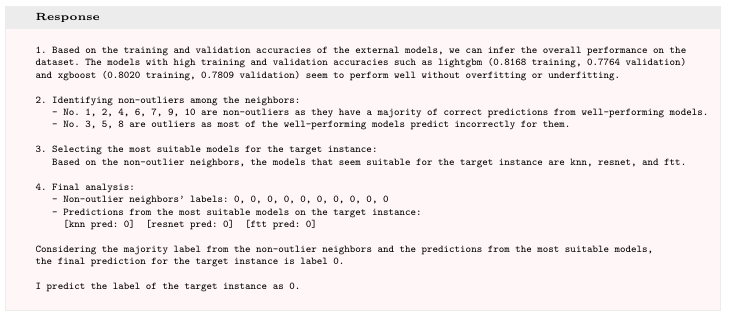}
    \caption{The response from gpt-3.5-turbo with~\name for~\autoref{fig:prompt}, where the true label
corresponding to the question is 0, and the model predicted 0, indicating a right prediction.}
    \label{lab:gpt3.5-turbo}
\end{figure}

\begin{figure}[h]
    \centering
    \includegraphics[width=\textwidth]{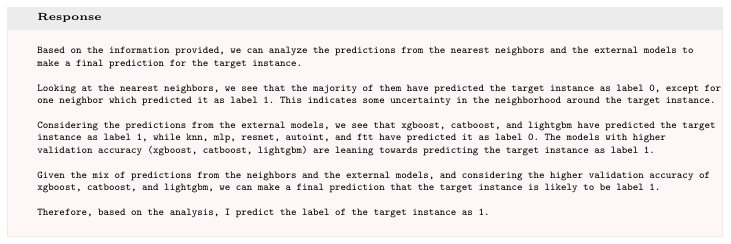}
    \caption{The response from gpt-3.5-turbo without~\name for~\autoref{fig:prompt}, where the true label
corresponding to the question is 0, and the model predicted 1, indicating a right prediction.}
    \label{lab:gpt3.5-turbo-without-cot}
\end{figure}

\begin{figure}[h]
    \centering
    \includegraphics[width=\textwidth]{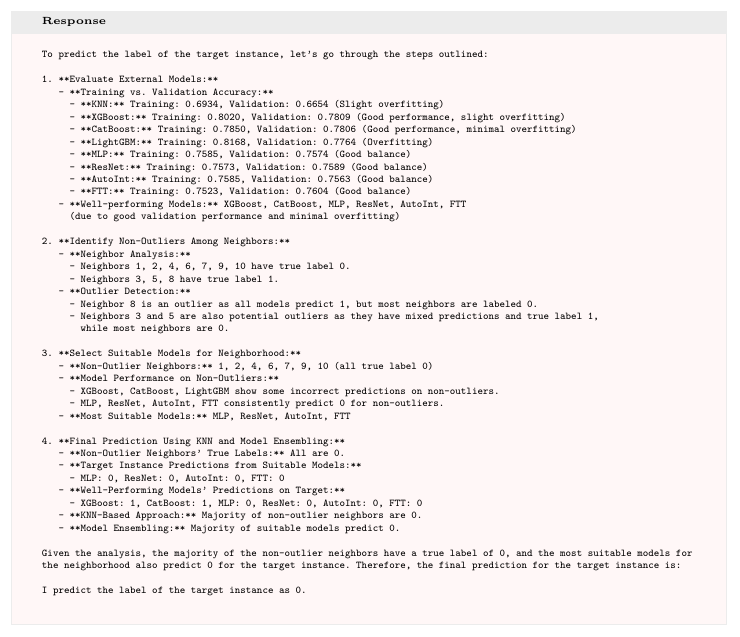}
    \caption{The response from gpt-4o for~\autoref{fig:prompt}, where the true label
corresponding to the question is 0, and the model predicted 0, indicating a right prediction.}
    \label{lab:gpt-4o}
\end{figure}

\begin{figure}[h]
    \centering
    \includegraphics[width=\textwidth]{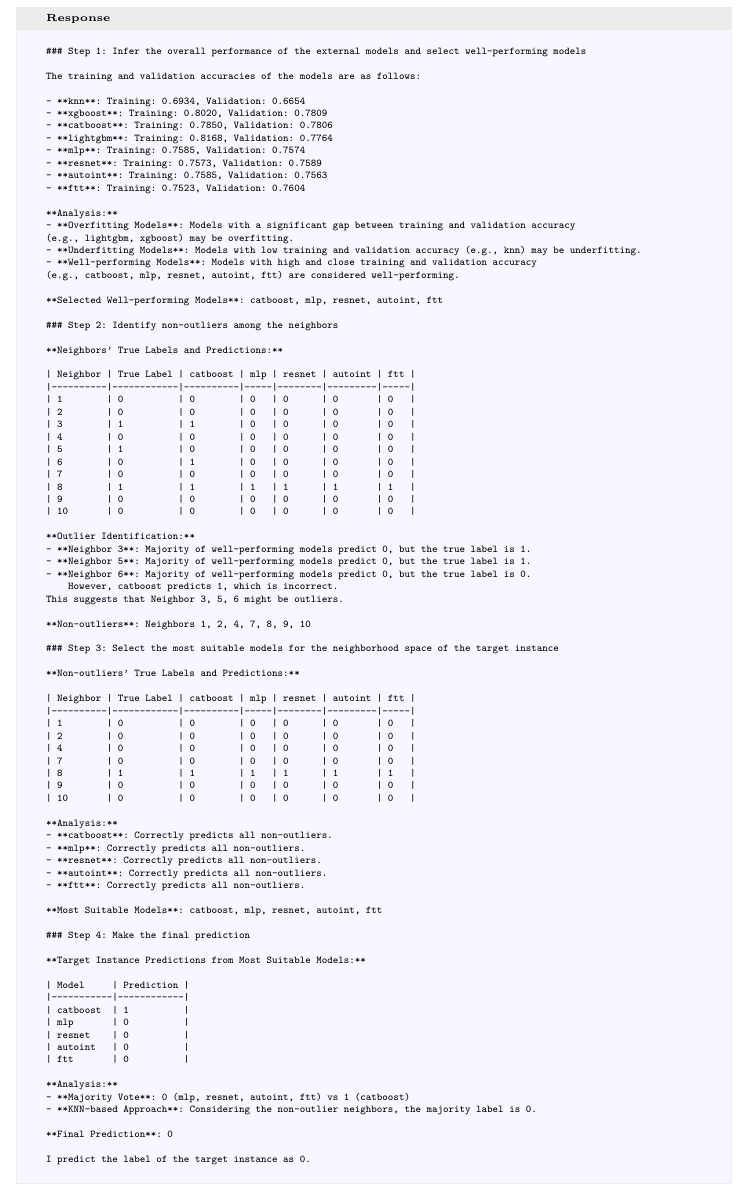}
    \caption{The response from DeepSeek-v3 for~\autoref{fig:prompt}, where the true label
corresponding to the question is 0, and the model predicted 0, indicating a right prediction.}
    \label{lab:DeepSeek-v3}
\end{figure}

\begin{figure}[h]
    \centering
    \includegraphics[width=\textwidth]{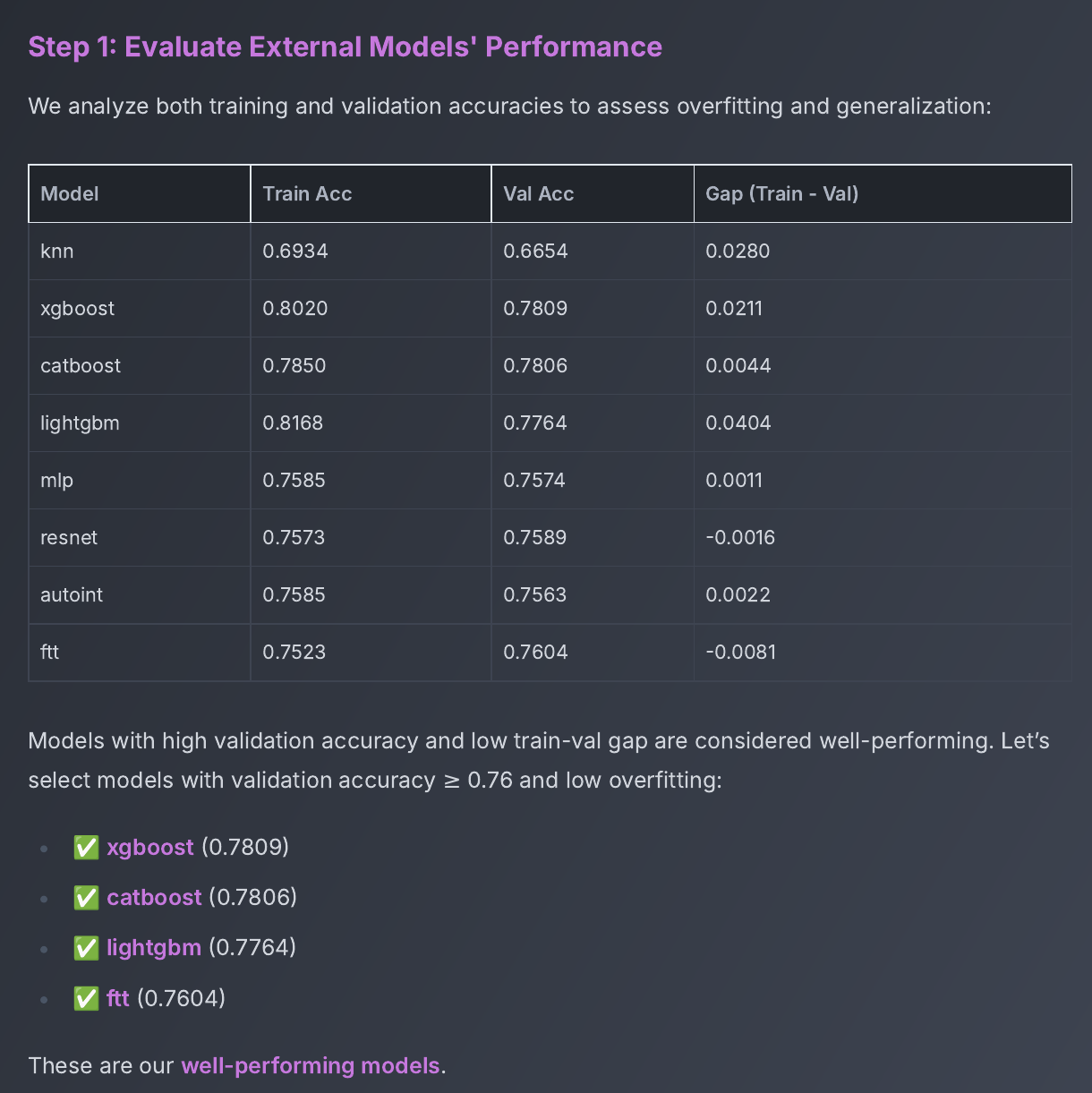}
    \caption{Step 1 of the response generated by the latest ChatGPT model for the example in~\autoref{fig:prompt}. This step demonstrates how the model begins to decompose the problem and reason through the input systematically.}
    \label{lab:step1}
\end{figure}

\begin{figure}[h]
    \centering
    \includegraphics[width=\textwidth]{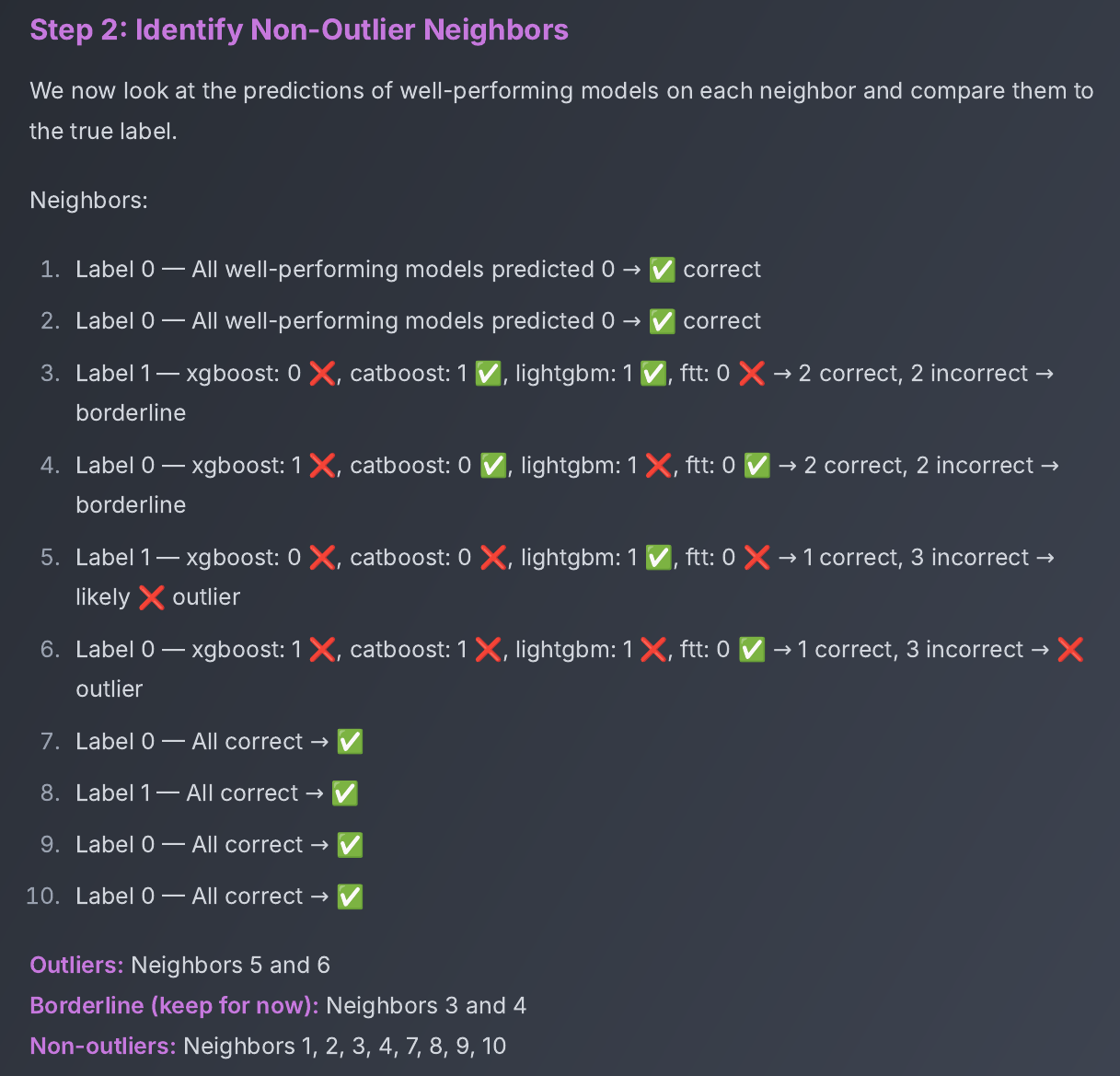}
    \caption{Step 2 of the response generated by the latest ChatGPT model for the example in~\autoref{fig:prompt}. }
    \label{lab:step2}
\end{figure}

\begin{figure}[h]
    \centering
    \includegraphics[width=\textwidth]{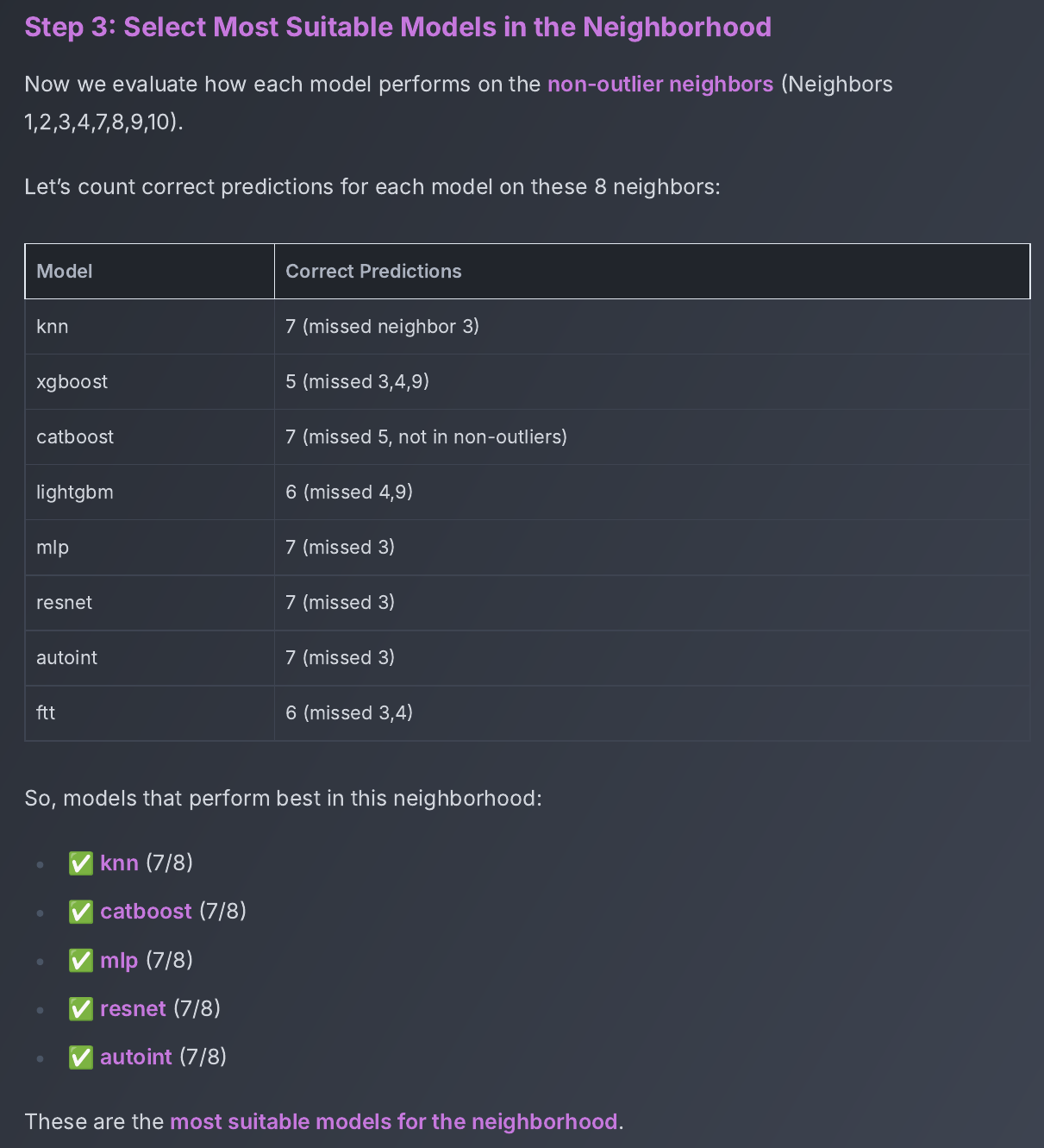}
    \caption{Step 3 of the response generated by the latest ChatGPT model for the example in~\autoref{fig:prompt}. }
    \label{lab:step3}
\end{figure}

\begin{figure}[h]
    \centering
    \includegraphics[width=\textwidth]{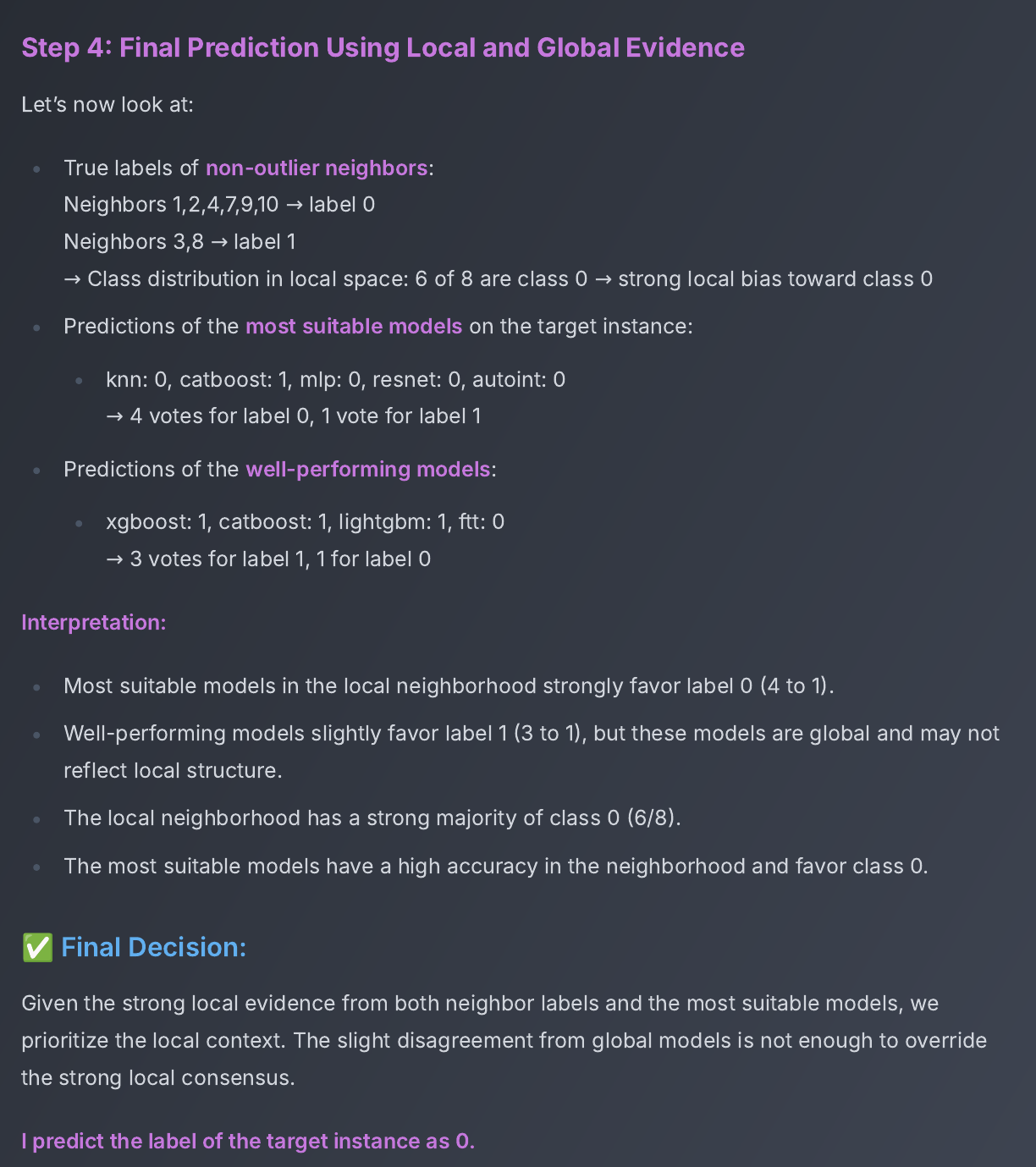}
    \caption{Step 4 of the response generated by the latest ChatGPT model for the example in~\autoref{fig:prompt}. }
    \label{lab:step4}
\end{figure}
\clearpage



\end{document}